\documentclass{article}

\usepackage{microtype}
\usepackage{graphicx}
\usepackage{subcaption}
\usepackage{booktabs}

\usepackage{hyperref}
\usepackage[T1]{fontenc}        
\usepackage[utf8]{inputenc}     
\usepackage[table]{xcolor}

\usepackage[preprint]{icml2026}

\usepackage{amsmath}
\usepackage{amssymb}
\usepackage{mathtools}
\usepackage{amsthm}

\usepackage[capitalize,noabbrev]{cleveref}

\theoremstyle{plain}

\theoremstyle{definition}

\theoremstyle{remark}

\usepackage{multirow}
\usepackage{stfloats}
\usepackage{placeins}
\usepackage{algorithm}
\usepackage{algorithmic}

\usepackage[disable,textsize=tiny]{todonotes}

\icmltitlerunning{OMTG: Towards One-to-Many Temporal Grounding}

\begin{document}

\twocolumn[
  \icmltitle{Towards One-to-Many Temporal Grounding}

  \icmlsetsymbol{equal}{*}

  \begin{icmlauthorlist}
    \icmlauthor{Qi Xu}{equal,WHU}
    \icmlauthor{Yue Tan}{equal,PKU}
    \icmlauthor{Shihao Chen}{WHU}
    \icmlauthor{Jiahao Meng}{PKU}
    \icmlauthor{Anna Wang}{NTU}
    \icmlauthor{Shunping Ji}{WHU}
    \icmlauthor{Hao Fei}{OX}
    \icmlauthor{Jason Li}{NTU}
  \end{icmlauthorlist}

  \icmlaffiliation{WHU}{Wuhan University}
  \icmlaffiliation{PKU}{Peking University}
    \icmlaffiliation{NTU}{Nanyang Technological University}
  \icmlaffiliation{OX}{University of Oxford}

  \icmlcorrespondingauthor{Jason Li}{xiangtai94@gmail.com}

  \icmlkeywords{Machine Learning, ICML, MLLM, Temporal Grounding}

  \centerline{Project Page: \url{https://insomniaaac.github.io/OMTG/}}

  \vskip 0.3in
]

\printAffiliationsAndNotice{}

\begin{abstract}
Temporal Grounding (TG) aims to localize video segments corresponding to a textual query. Prior research predominantly focuses on single-segment retrieval. Real-world scenarios, however, often require localizing multiple disjoint segments for a single query—a setting we term \textbf{One-to-Many Temporal Grounding (OMTG)}. 

Previous state-of-the-art MLLMs, optimized for one-to-one settings, struggle in this context, often yielding near-zero scores due to a lack of event cardinality perception. 

To bridge this gap, we present a systematic solution with three key contributions. First, we establish the first comprehensive OMTG benchmark, introducing Count Accuracy (C-Acc) and Effective Temporal F1 (EtF1) as evaluation metrics. 

Second, we curate a high-quality OMTG dataset comprising 56k samples through a sophisticated construction pipeline. 

Third, we develop novel temporal and caption reward functions specifically designed for OMTG. 

In particular, the caption reward leverages Chain-of-Thought reasoning over dense video captions to explicitly guide policy optimization toward both preciseness and completeness.

Extensive experiments show our model achieves a new state-of-the-art EtF1 of 43.65\% on OMTG Bench, outperforming Gemini 2.5 Pro and Seed-1.8 by 15.85\% and 15.61\%, respectively.
\end{abstract}

\section{Introduction}
\label{sec:intro}

\begin{figure}[ht]
    \centering
    \includegraphics[width=\linewidth]{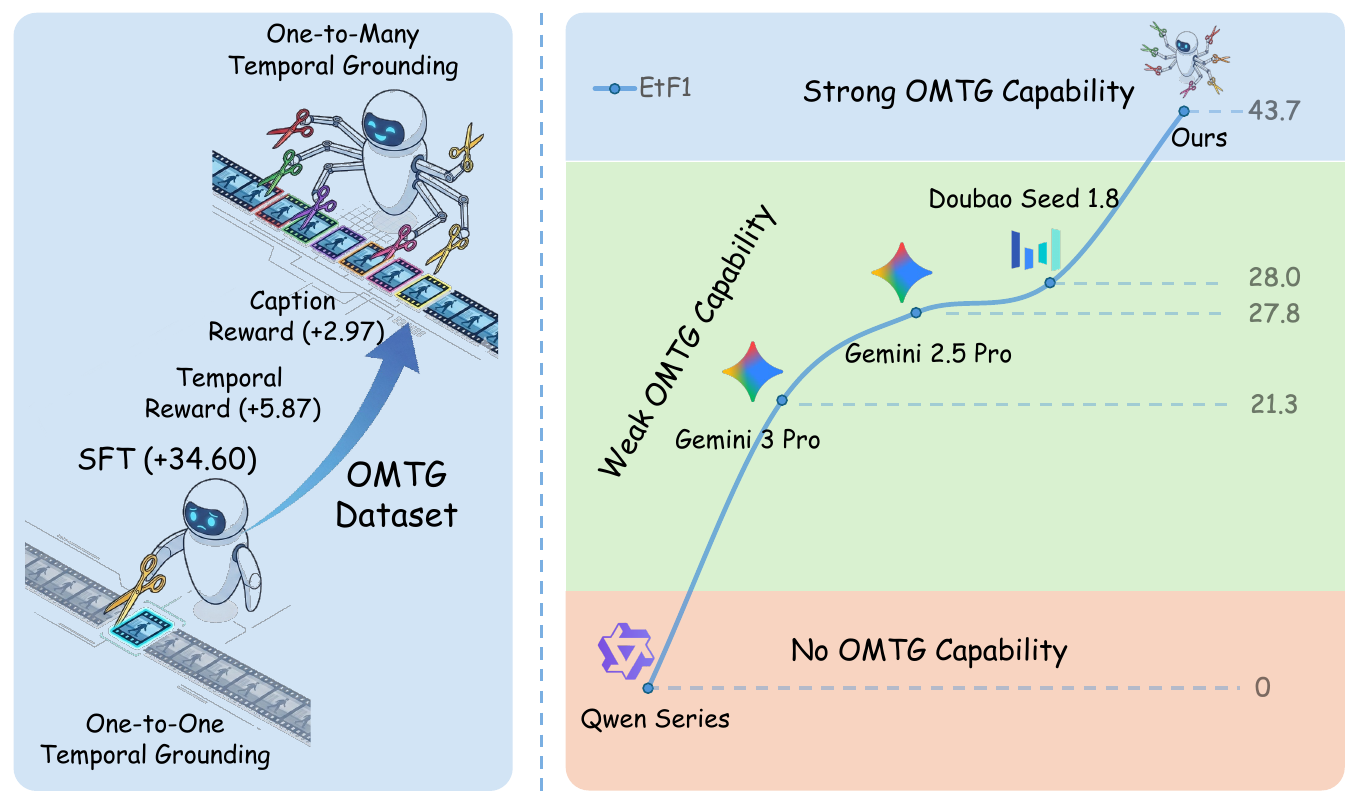}
    \caption{
\textbf{The Road to One-to-Many Temporal Grounding.} 
\textbf{Left:} Leveraging our OMTG dataset, we empower the model to evolve from One-to-One to One-to-Many through SFT and RL, underpinned by novel temporal and caption rewards.
\textbf{Right:} The capability landscape on the proposed OMTG Bench. 
}
    \label{fig:teaser}
    \vspace{-2em}
\end{figure}

Temporal Grounding aims to localize specific temporal segments within a video that semantically correspond to a given natural language query. 

As a fundamental task in video understanding, it has witnessed significant advancements driven by Multi-modal Large Language Models (MLLMs)~\cite{lin2023univtg,unitime,timer1,timesuite,timechat,videochat_r1}.

However, conventional research has predominantly focused on the one-to-one correspondence between queries and temporal segments. In real-world scenarios, video content is inherently dynamic and repetitive, with a single semantic action (e.g., \textit{"a person clapping"}) recurring at multiple distinct intervals. This characteristic gives rise to the One-to-Many Temporal Grounding problem, which requires identifying \textit{all} disjoint time segments semantically consistent with a query. Accurately retrieving the complete set of occurrences, rather than a single instance, is essential for a comprehensive understanding of complex video narratives.

To bridge this gap, we formally define \textbf{O}ne-to-\textbf{M}any \textbf{T}emporal \textbf{G}rounding (OMTG) as a set generation task within the MLLM framework. Recognizing that standard metrics for one-to-one grounding (e.g. tIoU, R@1) are ill-suited for this setting, we introduce a rigorous evaluation suite: Temporal F1-Score (tF1) to balance precision and recall, Count Accuracy (C-Acc) to assess event cardinality perception, and Effective Time F1 (EtF1) to strictly penalize incomplete retrieval and hallucinations. 

Furthermore, we establish the first comprehensive benchmark tailored for OMTG. Our extensive evaluation of state-of-the-art open-source and proprietary MLLMs, as illustrated in Figure \ref{fig:teaser}, reveals a critical capability gap: existing open-source models and traditional TG experts struggle significantly in the OMTG task, often yielding near-zero EtF1 scores, advanced proprietary models (e.g., Gemini series~\cite{gemini2_5}, Seed-1.8~\cite{seed1_8}) demonstrate weak-OMTG-capability, and our model significantly outperforms all baselines, reaching the strong-OMTG-capability zone. This stark contrast underscores the urgency of exploring this new direction.

To tackle these challenges, we devise a sophisticated data pipeline to construct 56k high-quality training samples. Leveraging this data, we propose a two-stage training strategy that synergizes Supervised Fine-Tuning (SFT) with subsequent Reinforcement Learning (RL). We employ two complementary rewards: caption rewards that leverages dense video captions with Chain-of-Thought reasoning to comprehend complex event structures, and temporal rewards that directly supervises temporal boundaries for precise localization. Notably, we observe that RL training on the OMTG task also improves standard one-to-one temporal grounding performance.

Extensive experiments demonstrate the superiority of our approach. Our model surpasses both leading open-source and proprietary models, achieving an EtF1 score of \textbf{43.65\%} on the OMTG Bench. This performance sets a new state-of-the-art, outperforming the previous best proprietary models Gemini 2.5 Pro and Seed-1.8 by significant margins of \textbf{15.85\%} and \textbf{15.61\%}, respectively.
\section{Related Work}
\label{sec:related_work}

\textbf{MLLMs for Video Temporal Grounding.}
MLLMs extend LLMs to visual modalities by unifying language, image, and video understanding within a single reasoning framework~\cite{llava, blip2, flamingo, videollama, videochat}. This task has also been reshaped by these models. Early TG methods relied on visual encoders with task-specific heads~\cite{man, qdln, umts}, while recent approaches directly leverage MLLMs' cross-modal reasoning capabilities via instruction tuning, causal event modeling, and hierarchical reasoning~\cite{huang2024vtimellm-internvid-vtime, timechat, trace, moment10m, videomind}. The training paradigms span supervised fine-tuning~\cite{sevila, llavamr}, reinforcement learning~\cite{timer1, videochat_r1}, and training-free methods~\cite{tfvtg, devi}. Despite these advances, existing methods largely inherit a \emph{one-to-one} supervision assumption, limiting their ability to handle complex real-world scenarios.

\textbf{Video Temporal Grounding Benchmarks and Datasets.}
Existing TG benchmarks suffer from annotation noise~\cite{charades-sta,qvhighlights,activitynet-captions, timelens} and, critically, a rigid \emph{one-to-one} formulation that fails to capture recurring or overlapping events. This limitation extends to training data: despite scaling efforts via MLLM generation~\cite{vidmorp, internvidg} or large-scale collection~\cite{moment10m, huang2024vtimellm-internvid-vtime}, current datasets rarely provide \emph{one-to-many} supervision. 

\textbf{Reinforcement Learning for Video MLLMs.}
Reinforcement learning has proven effective in improving the visual and cross-modal reasoning capabilities of MLLMs through verifiable or preference-based rewards~\cite{openai2023gpt4, zhou2025r1zero,zhan2025visionr1,deng2025curriculumrlvlm,liu2025visualrft,yang2025r1onevision,zhang2025r1vl}. More recently, RL has been applied to video MLLMs to better model spatio-temporal structure and long-range dependencies. Many methods~\cite{meng2025open,feng2025videor1,videochat_r1,videochat_r15,meng2025cyberv} introduce rule-based or perception-aware rewards and test-time scaling to significantly improve the model's temporal reasoning ability. However, existing RL-based approaches primarily optimize for a one-to-one temporal grounding task. In contrast, we present the sophisticated RL pipeline for One-to-Many Temporal Grounding. 

\section{One-to-Many Temporal Grounding}

\subsection{Problem Formulation}
\label{sec:problem_definition}

We formulate One-to-Many Temporal Grounding as a generative task under the MLLM framework. 
Given an input video $V = \{f_t\}_{t=1}^{T}$ consisting of $T$ visual frames and a textual query $Q = \{w_l\}_{l=1}^{L}$, the objective is to localize \emph{multiple} temporal segments in the video that correspond to repeated semantic occurrences of the query.

Specifically, we learn a mapping function $\mathcal{F}_{\theta}$, parameterized by an MLLM, which directly generates a natural language response:
\begin{equation}
    Y = \mathcal{F}_{\theta}(V, Q)
\end{equation}
The generated sequence $Y$ encodes a structured description of temporal intervals associated with the query events. 
A deterministic parsing function $\phi(\cdot)$ is applied to extract a set of predicted temporal segments
\begin{equation}
    \mathcal{P} = \phi(Y) = \{(\hat{s}_m, \hat{e}_m)\}_{m=1}^{M}
\end{equation}
where $\hat{s}_m$ and $\hat{e}_m$ denote the predicted start and end timestamps of the $m$-th instance, and $M$ is the number of predicted segments.

The ground-truth annotations are given as a set of temporal intervals
\begin{equation}
    \mathcal{G} = \{(s_k, e_k)\}_{k=1}^{K}
\end{equation}
where $K \geq 1$ denotes the number of semantic occurrences of the query in the video.
The learning objective is to generate a response $Y$ such that the extracted predictions $\mathcal{P}$ closely match $\mathcal{G}$ in terms of both cardinality (i.e., M = K) and temporal
boundaries.

\begin{figure*}[ht]
    \centering
    \includegraphics[width=\linewidth]{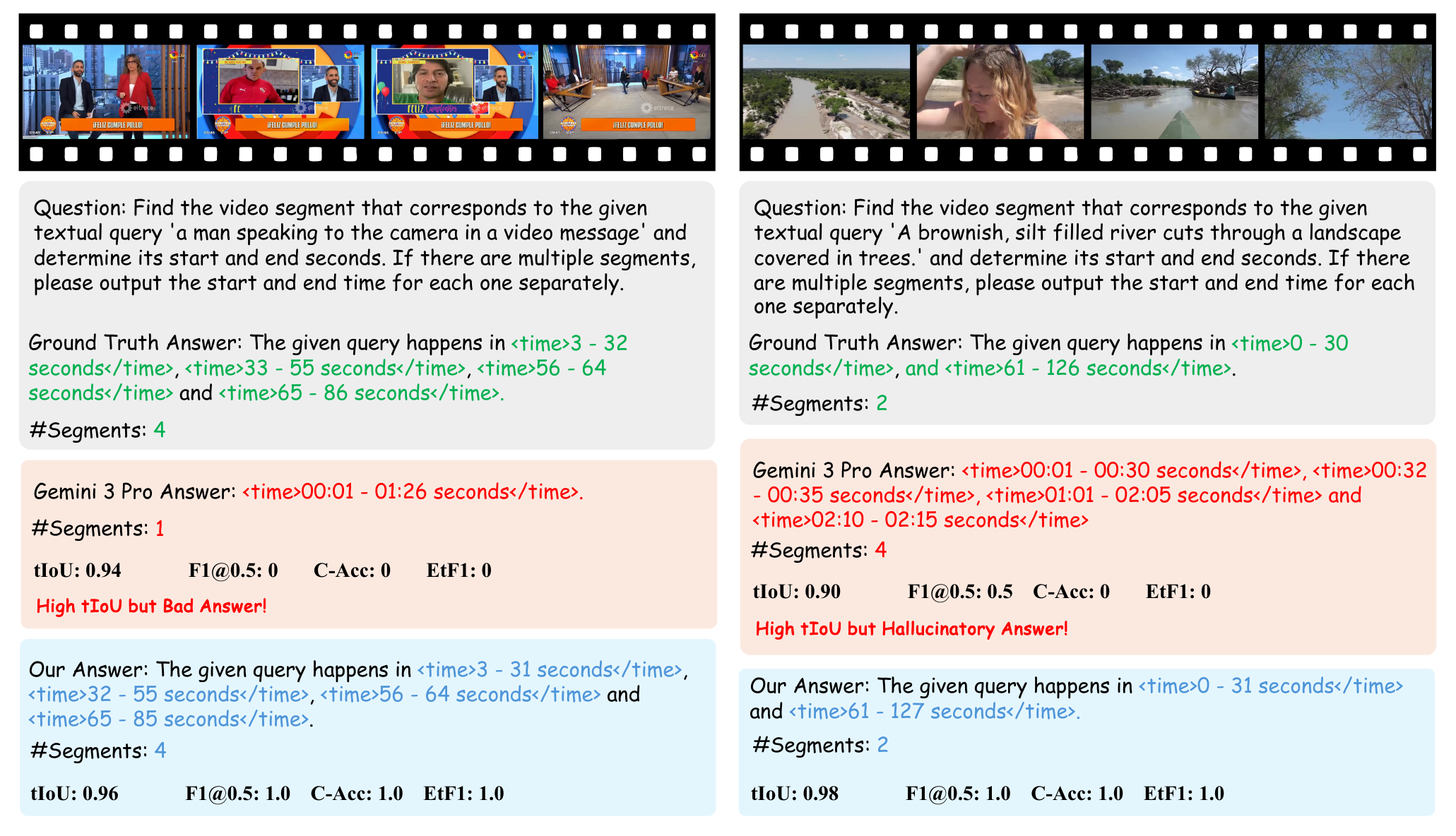}
    \caption{
    \textbf{The deceptiveness of tIoU in One-to-Many Temporal Grounding.}
    Gemini 3 pro achieves high tIoU ($>0.9$) in both examples despite event counting failures: under-segmentation (Left) and over-segmentation (Right). tIoU metrics fail to capture these counting errors. In contrast, our proposed EtF1 strictly penalizes count mismatches, providing a rigorous evaluation that highlights our method's superior precision in both event counting and localization.
    }
    \label{fig:metric_illustration}
\end{figure*}

\subsection{Metrics}
\label{sec:metric}

One-to-One Temporal Grounding benchmarks~\cite{charades-sta, activitynet-captions, qvhighlights, timelens} predominantly adopt Recall@1 (R@1) with a temporal IoU (tIoU) threshold as the evaluation metric. 
In the One-to-One setting, where each query is associated with a unique ground-truth segment, this metric is sufficient, as a correct retrieval simultaneously satisfies both precision and recall.

In contrast, One-to-Many Temporal Grounding poses fundamentally different evaluation challenges, since a single query corresponds to a \emph{set} of ground-truth segments. 
As illustrated in Fig.~\ref{fig:metric_illustration}, the distinction between precision and recall becomes non-trivial.
A high recall score only indicates that some ground-truth instances are covered, but fails to penalize redundant or hallucinated predictions. Moreover, the tIoU metric, while effective in measuring temporal overlap, is insensitive to the structural composition of events.
For example, if a model incorrectly merges two semantically distinct events separated by a short temporal gap into a single continuous segment, the resulting tIoU can remain deceptively high (e.g., 0.9) due to dominant overlap, despite the model failing to distinguish multiple occurrences and producing incorrect cardinality.

Therefore, a holistic evaluation of OMTG requires decoupling instance-level coverage and prediction correctness, explicitly measuring both precision and recall, and employing the F1-score to jointly assess temporal localization quality and instance-level fidelity. Furthermore, we introduce Effective Temporal F1-score which conditions F1-score on event cardinality to provides a rigorous evaluation of OMTG task.

We formulate the evaluation as a bipartite matching problem between the predicted segments $\mathcal{P}$ and the ground-truth segments $\mathcal{G}$. 
Given an IoU threshold $\xi$, we apply the Hungarian algorithm to compute the optimal one-to-one matching that maximizes the total temporal IoU. 
Based on this matching, we define the following evaluation metrics:

\textbf{Temporal IoU (tIoU).}
    To assess the overall temporal coverage between predictions and ground truth, we compute the Intersection over Union (IoU) between their temporal unions.
    Specifically, for each sample $i$, let $\cup \mathcal{P}_i$ and $\cup \mathcal{G}_i$ denote the unions of all predicted and ground-truth segments, respectively. The dataset-level tIoU is defined as
    \begin{equation}
        \text{tIoU} = \frac{1}{N} \sum_{i=1}^{N} 
        \frac{\text{length}\big((\cup \mathcal{P}_i) \cap (\cup \mathcal{G}_i)\big)}
        {\text{length}\big((\cup \mathcal{P}_i) \cup (\cup \mathcal{G}_i)\big)}
    \end{equation}
    where $\text{length}(\cdot)$ measures the total duration of a set of temporal intervals and $N$ denotes the number of samples.

\textbf{Temporal Precision and Recall.}
Based on the optimal bipartite matching under IoU threshold $\xi$, we define instance-level Temporal Precision and Temporal Recall to explicitly characterize prediction correctness and coverage in the One-to-Many setting.
Temporal Precision measures the fraction of predicted segments that correctly match ground-truth instances, penalizing redundant or hallucinated predictions, while Temporal Recall measures the fraction of ground-truth instances that are successfully localized.
Formally, for sample $i$, they are defined as
\begin{equation}
    tP_i@\xi = \frac{TP_i@\xi}{M_i}, \qquad
    tR_i@\xi = \frac{TP_i@\xi}{K_i}
\end{equation}
where $TP_i@\xi$ denotes the number of matched prediction–ground-truth pairs whose IoU exceeds $\xi$, $M_i$ is the number of predicted segments, and $K_i$ is the number of ground-truth segments.

\textbf{Temporal F1-Score (tF1).}
To jointly evaluate temporal precision and temporal recall in One-to-Many Temporal Grounding, we report the Temporal F1-Score at an IoU threshold $\xi$:
\begin{equation}
    tF1@\xi=\frac{1}{N}\sum_{i=1}^{N} 2 \cdot \frac{tP_{i}@\xi\cdot tR_{i}@\xi}{tP_{i}@\xi + tR_{i}@\xi}
\end{equation}

\textbf{Count Accuracy (C-Acc).} To explicitly evaluate the model's ability to perceive the correct number of event occurrences—a core challenge in One-to-Many Temporal Grounding—we introduce the Count Accuracy metric. It measures the percentage of test samples where the number of predicted segments exactly matches the number of ground truth segments.
\begin{equation}
    \text{C-Acc}=\frac{1}{N}\sum_{i=1}^{N}\mathbf{1}(M_i=K_i)
\end{equation}
where $N$ is the total number of samples in the dataset, $M_i$ is the predicted count, $K_i$ is the ground truth count, and $\mathbf{1}(\cdot)$ is the indicator function. A higher C-Acc indicates that the model has learned to count the occurrences of an event in the video.

\textbf{Effective Temporal F1-Score (EtF1).}
To jointly enforce accurate localization and correct instance counting, we propose the Effective Temporal F1-Score (EtF1), which conditions Temporal F1-Score on count consistency:
\begin{equation}
    \text{EtF1}=\frac{1}{N\cdot|\Xi|}\sum_{\xi\in\Xi}\sum_{i=1}^{N}
    \mathbf{1}(M_i=K_i)\cdot
    \frac{2 \cdot tP_{i}@\xi\cdot tR_{i}@\xi}{tP_{i}@\xi + tR_{i}@\xi}
\end{equation}
Here, the indicator $\mathbf{1}(M_i=K_i)$ acts as a gating function that assigns zero score to samples with incorrect predicted cardinality, and $\Xi=\{0.3,0.5,0.7\}$ denotes the set of IoU thresholds.
By explicitly coupling instance-level precision–recall with event-count correctness, EtF1 provides a rigorous and holistic evaluation of OMTG task.

\subsection{OMTG Benchmark}

We construct a high-quality benchmark consisting of 340 manually curated samples spanning diverse domains, including sports, cooking, and news. We randomly sampled and manually curated videos from the test sets of Charades~\cite{charades-sta}, ActivityNet~\cite{activitynet-captions}, QVHighlights~\cite{qvhighlights}, VTimeLLM~\cite{huang2024vtimellm-internvid-vtime}, and Moment10m~\cite{moment10m}, ensuring that the video sources of our benchmark have no overlap with the training set. Each sample is annotated with precise boundaries and verified by independent experts with a consistency rate exceeding 90\%.

The benchmark presents a diverse and challenging distribution. The number of ground truth segments per query ranges from 2 to 20; while the majority (62.2\%) involve 2-3 instances, a significant portion (15\%) contains more than 6 occurrences, posing a severe test for counting ability. Regarding temporal duration, the videos span from 21 seconds to over 17 minutes (avg. 221.6s), ensuring robust evaluation across both short clips and long-form narratives. Additional detailed statistics and examples of our benchmark are provided in the Appendix \ref{sec:benchmark_statistics}.

\section{Method}
\label{sec:method}
\subsection{Constructing High-Quality OMTG Dataset}

\begin{figure*}[ht]
    \centering
    \includegraphics[width=\textwidth]{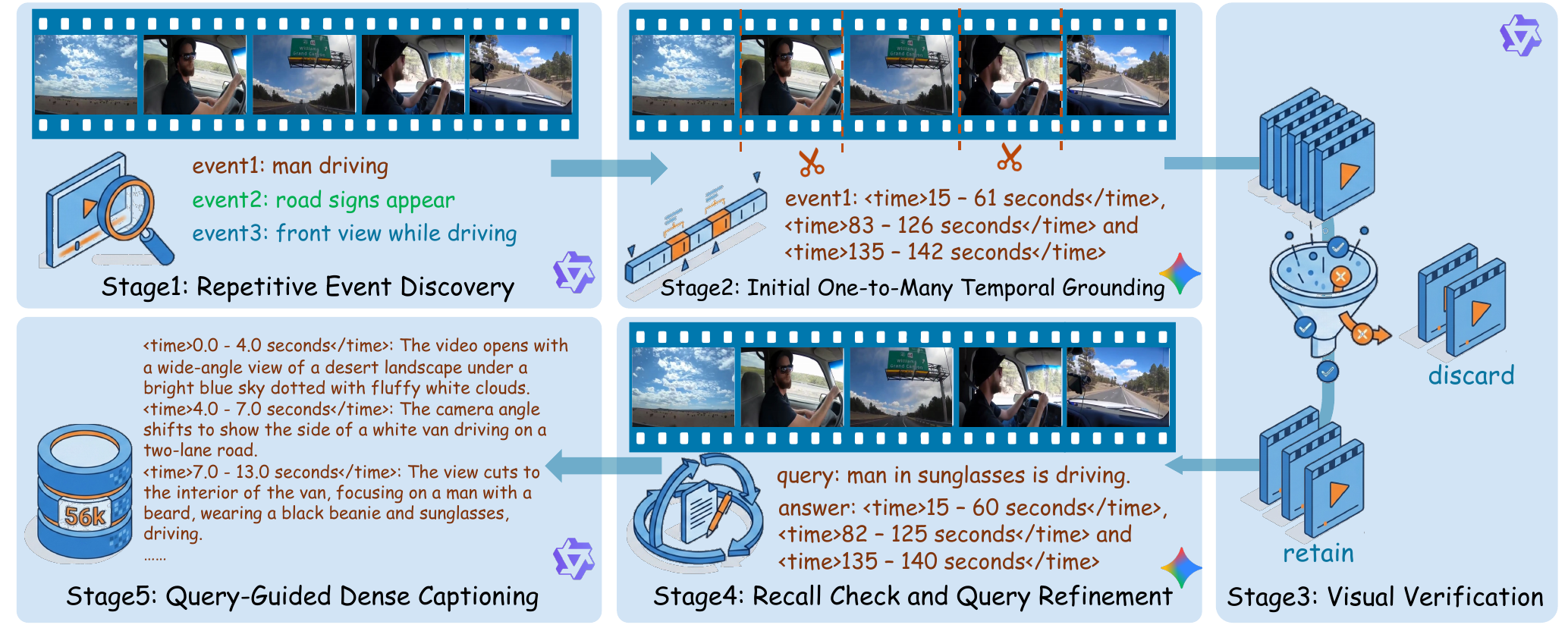}
    \caption{\textbf{Overview of our data construction pipeline:} The annotation pipeline includes repetitive event discovery, initial one-to-many grounding, strict visual verification, recall check and query refinement.}
    \label{fig:data_pipeline}
    \vspace{-1em}
\end{figure*}

\begin{figure}
    \centering
    \includegraphics[width=1.0\linewidth]{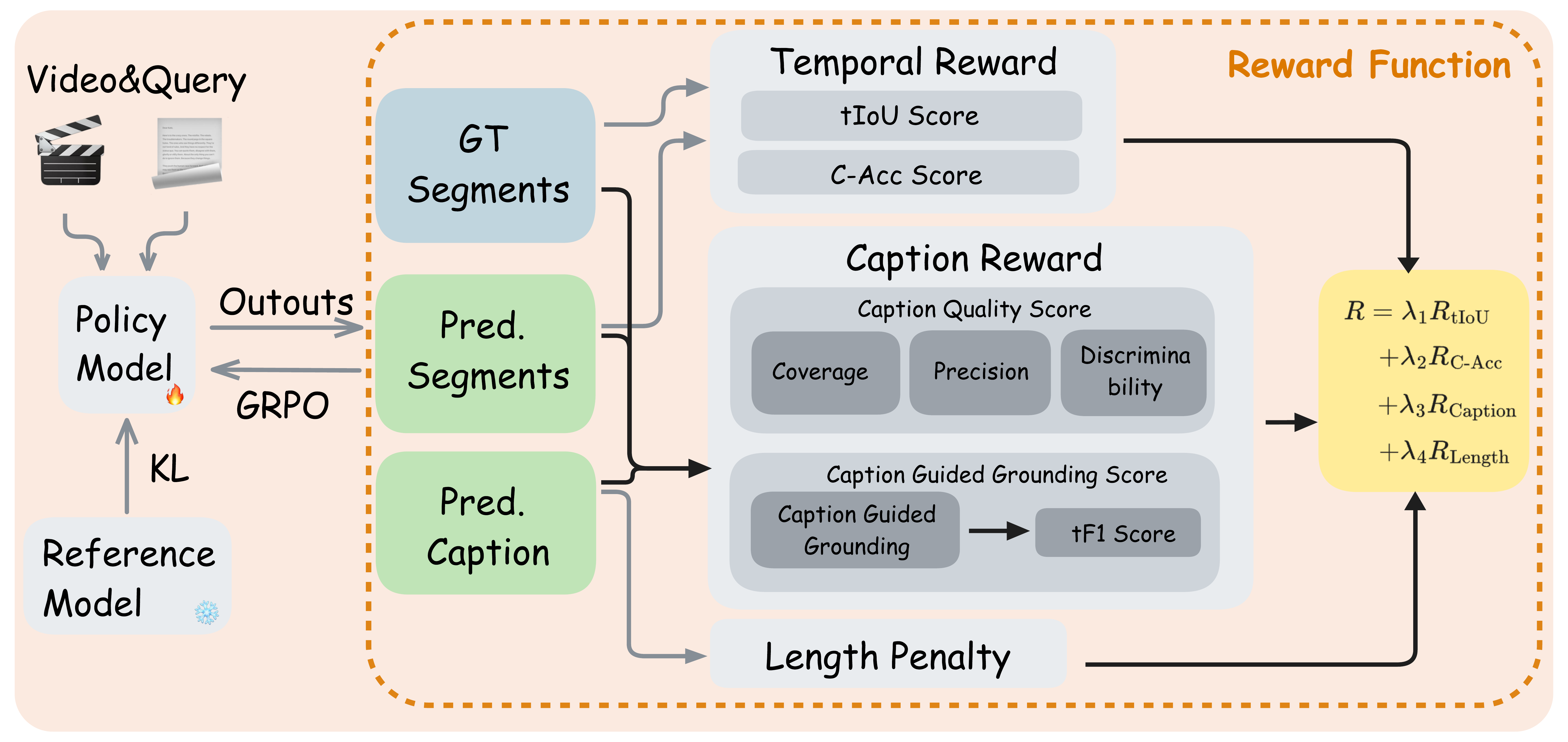}
    \caption{\textbf{Composite reward function optimized via GRPO.} The framework combines rule-based rewards for temporal precision with an LLM-as-a-judge mechanism for caption quality evaluation to improve one-to-many temporal grounding.}
    \label{fig:reward_function}
    \vspace{-12pt}
\end{figure}

To facilitate the training of robust OMTG models, we construct the \textbf{OMTG Dataset}, a high-quality instruction tuning dataset comprising approximately 56k samples. The raw videos are sourced from diverse public datasets, including Cosmos-Cap\cite{wang2024cosmo_cap}, Moment-10M\cite{moment10m}, and VTimeLLM\cite{huang2024vtimellm-internvid-vtime}. As shown in Fig \ref{fig:data_pipeline}, to transform these raw videos into precise one-to-many supervision signals, we design a rigorous four-stage automated pipeline leveraging state-of-the-art MLLMs.

\textbf{Stage 1: Repetitive Event Discovery.}
We employ the powerful \textit{Qwen3-VL-235B} model as the event discoverer. The model scans the raw videos to identify salient events that occur multiple times. For each identified repetitive event, the model generates a descriptive query, serving as the initial prompt for the subsequent stages.

\textbf{Stage 2: Initial One-to-Many Grounding.}
Using the generated queries, we prompt \textit{Gemini 2.5 Pro} to perform fine-grained temporal grounding. The model is instructed to scan the video and return a set of precise start and end timestamps for all occurrences of the event. This step transforms the semantic query into preliminary temporal annotations.

\textbf{Stage 3: Strict Visual Verification.}
To eliminate hallucinations and inaccurate boundaries, we implement a strict visual verification protocol. We temporally crop the video segments based on the timestamps from Stage 2. Each cropped clip is then fed back into \textit{Qwen3-VL-235B} to verify whether the visual content strictly aligns with the textual query. We adopt an \textit{"All-or-Nothing"} filtering strategy: if \textbf{any} single segment within a sample fails the verification (i.e., the model judges it as a mismatch), the entire data sample is discarded. This rigorous filtering ensures that the final dataset maintains an exceptionally high precision rate.

\textbf{Stage 4: Recall Check and Query Refinement.}
Finally, the surviving samples undergo a semantic refinement phase using \textit{Gemini 2.5 Pro}. We feed the video, the query, and the verified timestamps back to the model to perform a dual check: (1) \textbf{Recall Check}: Identifying if any valid segments were missed in the previous stages; (2) \textbf{Query Refinement}: Polishing the query text to ensure it is unambiguous and accurately describes the visual commonality of all segments.

\textbf{Stage 5: Query-Guided Dense Captioning.} Finally, we generate comprehensive, fine-grained captions using \textit{Qwen3-VL-235B}. The model is prompted to identify \textbf{all} distinct activity events in the video. Crucially, the refined queries from Stage 4 serve as mandatory guidance: the model must incorporate their information through detailed elaboration. This yields dense, semantically precise captions that contextualize the repetitive events within the full activity stream. Leveraging these captions, we construct Chain-of-Thought (CoT) and design a caption reward upon this foundation to better guide policy optimization and enhance temporal grounding accuracy.

This pipeline results in \textbf{56k} high-fidelity training samples with dense, verified annotations. We split the dataset into 46k samples for SFT training and 10k samples for RL training. Detailed prompts and additional implementation details are provided in the Appendix~\ref{sec:appendix_datapipeline}.

\subsection{Achieving Preciseness and Completeness OMTG}
\begin{table*}[h!]
\caption{\textbf{Main results on the OMTG Bench.} We conduct a comprehensive assessment of representative open-source and proprietary MLLMs to establish a comprehensive baseline for the OMTG task. Metrics include Count Accuracy (C-Acc), Temporal F1-Scores (tF1@0.3/0.5/0.7), average temporal IoU (tIoU), and Effective Temporal F1-Score (EtF1) are reported. \textbf{The benchmark reveals a critical capability gap:} standard open-source models (e.g., Qwen2.5-VL series) yield \textbf{0\% C-Acc}, failing to capture the one-to-many complexity.}
\centering
\small
\renewcommand{\arraystretch}{1.2}
\resizebox{0.90\textwidth}{!}{ 
\begin{tabular}{
  >{\raggedright\arraybackslash}p{0.33\linewidth}  
  >{\centering\arraybackslash}p{0.08\linewidth}     
  >{\centering\arraybackslash}p{0.08\linewidth}    
  >{\centering\arraybackslash}p{0.08\linewidth}    
  >{\centering\arraybackslash}p{0.08\linewidth}    
  >{\centering\arraybackslash}p{0.08\linewidth}    
  >{\centering\arraybackslash}p{0.08\linewidth}    
}
\toprule
\textbf{Model} & \textbf{C-Acc} & \textbf{tF1@0.3} & \textbf{tF1@0.5} & \textbf{tF1@0.7} & \textbf{tIoU} & {\textbf{EtF1}} \\ \midrule

Seed-1.8~\cite{seed1_8} & 38.12 & 67.13 & 54.67 & 38.79 & 56.81 & {28.04} \\
Gemini-2.5-Pro~\cite{gemini2_5} & 50.94 & 55.72 & 43.57 & 27.97 & 43.24 & {27.80} \\ 
Gemini-3-Pro~\cite{gemini2_5} & 30.63 & 58.30 & 47.75 & 29.89 & 47.63 & {21.30} \\ \midrule

Qwen2.5-VL-3B~\cite{qwen2-5-vl} & 0.00 & 15.17 & 7.01 & 2.86 & 11.60 & {0.00} \\
Qwen2.5-VL-7B~\cite{qwen2-5-vl} & 0.00 & 21.04 & 12.08 & 7.14 & 20.35 & {0.00} \\
Qwen2.5-VL-32B~\cite{qwen2-5-vl} & 0.00 & 16.81 & 9.66 & 4.76 & 18.32 & {0.00} \\
Qwen2.5-VL-72B~\cite{qwen2-5-vl} & 0.00 & 21.16 & 12.20 & 6.88 & 20.02 & {0.00} \\

Qwen3-VL-4B~\cite{Qwen3-VL} & 0.31 & 37.07 & 26.75 & 17.93 & 30.42 & {0.21} \\
Qwen3-VL-8B~\cite{Qwen3-VL} & 0.00 & 37.73 & 27.02 & 18.70 & 30.62 & {0.00} \\
Qwen3-VL-30B~\cite{Qwen3-VL} & 0.00 & 37.03 & 25.98 & 17.52 & 32.36 & {0.00} \\
Qwen3-VL-235B~\cite{Qwen3-VL} & 0.31 & 34.66 & 25.25 & 16.45 & 25.56 & {0.21} \\
VideoChat-R1-7B~\cite{videochat_r1} & 0.00 & 32.07 & 19.70 & 10.42 & 24.93 & {0.00} \\
VideoChat-R1.5-7B~\cite{videochat_r15} & 0.31 & 28.41 & 15.53 & 9.85 & 27.96 & {0.10} \\
Time-R1-7B~\cite{timer1} & 0.00 & 28.94 & 18.73 & 10.00 & 24.11 & {0.00} \\
UniTime~\cite{unitime} & 0.00 & 35.27 & 30.15 & 23.58 & 37.12 & {0.00} \\
Timelens-8B~\cite{timelens} & 0.00 & 39.14 & 32.76 & 22.58 & 32.38 & {0.00} \\
\rowcolor[HTML]{E8F0FB} 
\hline
\textbf{OMTG-4B} & \textbf{55.63} & \textbf{73.46} & \textbf{65.40} & \textbf{48.96} & \textbf{61.24} & {\textbf{43.65}} \\
\bottomrule
\end{tabular}
}

\label{tab:omtg_main}
\vspace{-0.5em}
\end{table*}

\begin{table*}[h!]
\centering
\begin{minipage}[t]{0.48\textwidth}
\small
\setlength{\tabcolsep}{2.8pt}  
\renewcommand{\arraystretch}{1.0}
\caption{\textbf{Performance gain of our method on OMTG Bench.}}
\vspace{-0.5em}

\resizebox{\textwidth}{!}{ 
\begin{tabular}{lcccccc}
\toprule
\textbf{Model} & \textbf{C-Acc} & \textbf{tF1@0.3} & \textbf{tF1@0.5} & \textbf{tF1@0.7} & \textbf{tIoU} & \textbf{EtF1} \\ \midrule
Base & 0.31 & 37.07 & 26.75 & 17.93 & 30.42 & 0.21 \\ \midrule
SFT & 44.06 & 69.57 & 61.23 & 45.63 & 56.94 & 34.81 \\
RL  & \textbf{55.63} & \textbf{73.46} & \textbf{65.40} & \textbf{48.96} & \textbf{61.24} & \textbf{43.65} \\
\bottomrule
\end{tabular}
}
\label{tab:4b_performance_gain}
\end{minipage}
\hfill 
\begin{minipage}[t]{0.50\textwidth}
\small
\setlength{\tabcolsep}{2.8pt} 
\renewcommand{\arraystretch}{1.0}
\caption{\textbf{Ablation on different reward functions on OMTG Bench.}}
\resizebox{\textwidth}{!}{ 
\begin{tabular}{lcccccc}
\toprule
\textbf{Reward Functions} & \textbf{C-Acc} & \textbf{tF1@0.3} & \textbf{tF1@0.5} & \textbf{tF1@0.7} & \textbf{tIoU} & \textbf{EtF1} \\ \midrule
$R_{\text{tIoU}}$ & +0.31 & +2.01 & +1.82 & +0.28 & +2.61 & +0.74 \\ 
$R_{\text{tIoU}}$ + $R_{\text{C-Acc}}$ & +7.50 & +4.69 & +3.86 & +3.27 & +4.03 & +5.87 \\ 
\rowcolor[HTML]{E8F0FB} 
$R_{\text{tIoU}}$ + $R_{\text{C-Acc}}$ + $R_{\text{Caption}}$ & \textbf{+11.57} & \textbf{+3.89} & \textbf{+4.17} & \textbf{+3.33} & \textbf{+4.30} & \textbf{+8.84} \\
\bottomrule
\end{tabular}
}
\label{tab:ablation_reward}
\end{minipage}
\vspace{-1.5em}
\end{table*}

We conduct SFT based on the OMTG Dataset. SFT stage facilitates the integration of fine-grained temporal localization details into the CoT reasoning process with dense video caption, from which the model can deduce the final grounding results.  

While SFT provides a strong initialization with chain-of-thought reasoning capabilities, including generating descriptive captions before final predictions, it often struggles to balance the trade-offs between retrieval completeness and localization precision. To address this, we design a composite reward function optimized via the GRPO~\cite{grpo} algorithm:
\begin{equation}
    R = \lambda_1 R_{\text{tIoU}} + \lambda_2 R_{\text{C-Acc}} + \lambda_3 R_{\text{Caption}} + \lambda_4 R_{\text{Length}}
    \label{eq:reward_function}
\end{equation}

where $R_{\text{tIoU}}$ and $R_{\text{C-Acc}}$ are defined following the metric formulations in Section~\ref{sec:problem_definition}. We set $\lambda_1 = \lambda_2 = \lambda_3 = 0.5$ and $\lambda_4 = -0.3$ to balance temporal localization quality, counting completeness, caption quality, and response conciseness. Figure~\ref{fig:reward_function} intuitively presents the overall design of our proposed reward function.

\paragraph{Temporal Reward.}
In Eq.~\ref{eq:reward_function}, $R_{\text{tIoU}}$ serves as a foundational component for refining temporal boundaries, whose effectiveness has been extensively validated in prior works~\cite{timer1,videochat_r1,videochat_flash,videochat_r15}. Complementing this, we introduce $R_{\text{C-Acc}}$ as a strict constraint that activates only when the predicted segment count exactly matches the ground truth, explicitly correcting the model's perception of event cardinality.

\paragraph{Caption Reward.} To encourage the model to generate informative intermediate reasoning during chain-of-thought, we introduce a caption quality reward $R_{\text{Caption}}$ that evaluates the descriptive captions produced before final predictions. We employ a two-part LLM-as-Judge evaluation framework using Qwen3-30B-A3B~\cite{yang2025qwen3} as the reward model.

The first part computes a \textit{Caption Quality Score} ($S_{\text{cq}}$), which evaluates three dimensions with access to ground truth: \textit{coverage} ($S_{\text{cov}}$), measuring whether all ground truth segments are matched; \textit{precision} ($S_{\text{prec}}$), assessing boundary alignment accuracy; and \textit{discriminability} ($S_{\text{disc}}$), determining whether captions provide unique contextual information:
\begin{equation}
S_{\text{cq}} = \mu_1 \cdot S_{\text{cov}} + \mu_2 \cdot S_{\text{prec}} + \mu_3 \cdot S_{\text{disc}}
\end{equation}

The second part computes a \textit{Caption Guided Grounding Score} ($S_{\text{cgg}}$), where the judge attempts to localize event timestamps by reading only the generated captions without access to the video. The predicted intervals are then compared against ground truth using tF1 scores to measure localization accuracy. This ensures that captions contain sufficient semantic information to independently support event localization.

The final caption reward combines these two components:
\begin{equation}
    R_{\text{Caption}} = \alpha \cdot S_{\text{cq}} + (1-\alpha) \cdot S_{\text{cgg}}
\end{equation}

Detailed formulations, coefficient settings and prompt templates are provided in Appendix~\ref{sec:appendix_reward}.

\paragraph{Length Penalty.} Excessively long captions introduce irrelevant details, dilute query-relevant temporal cues, and degrade localization performance. We therefore adopt a soft length penalty 
$R_{\text{Length}}$ that progressively penalizes responses exceeding predefined thresholds, preventing the model from being distracted from the core temporal grounding task. The specific formulation of the length penalty is detailed in the Appendix~\ref{sec:appendix_reward}.

We also explored alternative reward combinations such as $R_{\text{tIoU}} + R_{\text{C-Acc}}$. Through comprehensive ablation studies (see Section~\ref{sec:ablation}), we find that the combination of $R_{\text{tIoU}} + R_{\text{C-Acc}} + R_{\text{Caption}} + R_{\text{Length}}$ achieves strong performance for One-to-Many Temporal Grounding. By jointly optimizing these components via GRPO, we achieve a holistic alignment that ensures both accurate event counting and precise temporal localization.

\begin{table*}[ht]
\centering
\small
\setlength{\tabcolsep}{1.5pt}  
\renewcommand{\arraystretch}{1.1}  
\caption{\textbf{Main results on One-to-One Temporal Grounding (OOTG) benchmarks.} We benchmark the performance of various state-of-the-art proprietary and open-source models on TimeLens-Bench.}
\resizebox{\linewidth}{!}{  
\begin{tabular}{lcccc|cccc|cccc}
\toprule
\multirow{2}{*}{\textbf{Model}} & \multicolumn{4}{c}{\textbf{Charades}} & \multicolumn{4}{c}{\textbf{ActivityNet}} & \multicolumn{4}{c}{\textbf{QVHighlights}} \\
& R1@0.3 & R1@0.5 & R1@0.7 & tIoU & R1@0.3 & R1@0.5 & R1@0.7 & tIoU & R1@0.3 & R1@0.5 & R1@0.7 & tIoU \\
\midrule
\multicolumn{13}{l}{\textit{Proprietary Models}} \\
\rowcolor[HTML]{F0F0F0} 
GPT-4o~\cite{openai2023gpt4}  & 60.6 & 44.5 & 23.5 & 41.8 & 55.2 & 41.4 & 25.8 & 40.4 & 69.0 & 54.8 & 38.5 & 52.1 \\
\rowcolor[HTML]{F0F0F0} 
GPT-5~\cite{openai2023gpt4}  & 59.3 & 42.0 & 22.0 & 40.5 & 57.4 & 44.9 & 30.4 & 42.9 & 72.4 & 60.4 & 46.4 & 56.8 \\
\rowcolor[HTML]{F0F0F0} 
Gemini-2.5-Pro~\cite{gemini2_5}  & \underline{74.1} & \underline{61.1} & \underline{34.0} & \underline{52.8} & \underline{72.3} & \underline{64.2} & \underline{47.1} & \underline{58.1} & \underline{84.1} & \underline{75.9} & \underline{61.1} & \underline{70.4} \\
\midrule
\multicolumn{13}{l}{\textit{Open-Source Models}} \\
VideoChat-Flash-7B~\cite{videochat_flash}  & 60.2 & 37.9 & 17.8 & 39.7 & 35.5 & 21.8 & 10.5 & 24.8 & 45.2 & 30.6 & 16.7 & 32.7 \\
VideoChat-R1-7B~\cite{videochat_r1}  & 51.9 & 30.8 & 11.7 & 33.7 & 35.0 & 23.9 & 11.3 & 25.0 & 29.3 & 19.1 & 9.4 & 21.5 \\
Time-R1-7B~\cite{timer1}  & 57.9 & 32.0 & 16.9 & 36.6 & 44.8 & 31.0 & 19.0 & 33.1 & 65.8 & 51.5 & 36.1 & 49.2 \\
Qwen2.5-VL-7B ~\cite{qwen2-5-vl} & 59.7 & 37.8 & 16.6 & 39.3 & 44.1 & 31.0 & 16.1 & 31.4 & 41.5 & 27.8 & 15.2 & 31.6 \\
TimeLens-7B~\cite{timelens} & 70.5 & 55.6 & 28.4 & 48.8 & 62.8 & 51.0 & 32.6 & 46.2 & 74.1 & 62.7 & 43.1 & 56.0 \\
Qwen3-VL-4B~\cite{Qwen3-VL} & 67.1 & 47.5 & 24.1 & 45.6 & 62.9 & 51.0 & 34.3 & 47.2 & 77.6 & 66.0 & 50.9 & 61.7 \\
Qwen3-VL-8B~\cite{Qwen3-VL}  & 69.2 & 53.4 & 27.5 & 48.3 & 62.1 & 51.2 & 34.4 & 46.8 & 74.2 & 64.6 & 49.3 & 59.4 \\
\hline

OMTG-4B (SFT) & 70.4 & 56.9 & 31.2 & 50.0 & 59.5 & 50.2 & 34.4 & 45.8 & 76.5 & 65.4 & 50.8 & 61.5 \\
\rowcolor[HTML]{E8F0FB} 
OMTG-4B (RL) & \textbf{72.0} & \textbf{58.3} & \textbf{32.0} & \textbf{50.5} & \textbf{64.7} & \textbf{54.6} & \textbf{37.0} & \textbf{46.8} & \textbf{80.9} & \textbf{70.4} & \textbf{53.5} & \textbf{62.2} \\
\bottomrule
\end{tabular}
}

\label{tab:ootg_main}
\vspace{-2em}
\end{table*}

\section{Experiment}
\label{sec:exp:}

\subsection{Experiments Setup}
\textbf{Datasets and Metrics.}
We conduct a comprehensive evaluation across two distinct settings: One-to-Many Temporal Grounding (OMTG) and One-to-One Temporal Grounding (OOTG). For the OMTG task, we evaluate models on our proposed OMTG Bench. We report a holistic set of metrics, including Count Accuracy (C-Acc), Time F1-Score (tF1@0.3, 0.5, 0.7), our proposed Effective Time F1 (EtF1), and the traditional tIoU. For the OOTG task, we evaluate on the refined version of the TimeLens~\cite{timelens} dataset. Following standard protocols, we report Recall@1 (R@1) at IoU thresholds of 0.3, 0.5, and 0.7, alongside tIoU.

For training, we construct task-specific data mixtures. In the SFT stage, we utilize a high-quality mixture comprising 46k samples from our OMG-TG Dataset and 32k samples from TimeLens-100k~\cite{timelens}. In the subsequent RL stage, we exclusively utilize a subset of 10k samples from the OMG-TG Dataset to focus on alignment with complex grounding objectives.

\textbf{Implementation Details.}
Our primary experiments use Qwen3-VL-4B~\cite{Qwen3-VL} as the backbone.
During the SFT stage, the model is trained for 1 epoch on 16 NVIDIA H100 GPUs ($\sim$5 hours) using the AdamW\cite{adamw} optimizer. The learning rate is set to 1e-5 with a cosine scheduler and a 0.03 linear warmup. We use a global batch size of 64 with 4 gradient accumulation steps.

During the RL stage, we employ Group Relative Policy Optimization (GRPO) for 308 steps on 16 NVIDIA H100 GPUs ($\sim$30 hours). 

We perform 8 rollouts per prompt with a global batch size of 64. DeepSpeed~\cite{deepspeed} ZeRO-2 and Flash Attention~\cite{flashattention} are utilized for optimization.

\begin{figure*}[h!]
    \centering
    \includegraphics[width=\textwidth]{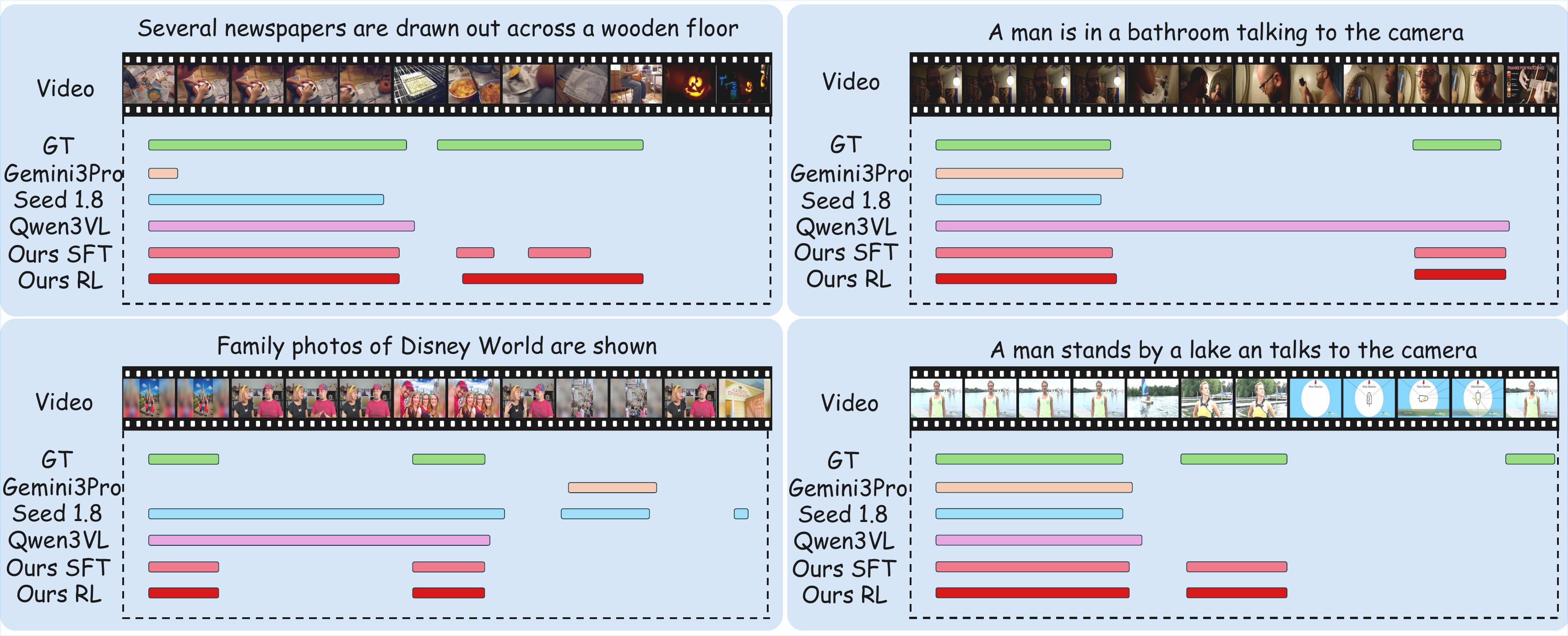}
    \caption{
\textbf{Qualitative visualization of One-to-Many Temporal Grounding.} 
We compare the predicted temporal segments from our model against state-of-the-art baselines across four diverse datasets. The green bars denote the Ground Truth. \textbf{Baselines:} Existing MLLMs (e.g., Gemini 3 Pro, Qwen3-VL) typically fail to capture the repetitiveness nature of events. They often retrieve only a single segment (e.g., Gemini 3 Pro in the top-left) or incorrectly merge distinct segments into a continuous span (e.g., Qwen3-VL, Seed-1.8).
\textbf{Ours:} In contrast, our model (Ours RL) accurately localizes all disjoint event occurrences, demonstrating superior capability in both event cardinality perception and boundary precision.
}
    \label{fig:qualitative}
    \vspace{-1em}
\end{figure*}

\subsection{Main Results}

\textbf{Results on OMTG Task.}
As shown in Table~\ref{tab:omtg_main}, we conduct a comprehensive evaluation on our proposed OMTG Bench. The results reveal a critical capability gap in existing MLLMs: open-source models and traditional TG experts struggle significantly in the OMTG task, often yielding near-zero EtF1 scores; advanced proprietary models (e.g., Gemini series~\cite{gemini2_5}, Seed-1.8~\cite{seed1_8}) demonstrate weak OMTG capability. In contrast, our OMTG-4B achieves state-of-the-art performance across all metrics, attaining an EtF1 of 43.65, which outperforms the best proprietary baselines by over 15.61\%. 

Qualitative comparisons in Figure~\ref{fig:qualitative} further illustrate the typical failure modes of existing models, including under-segmentation and over-segmentation, while demonstrating our method's superior ability to accurately identify all event occurrences with precise temporal boundaries.

Qualitative comparisons in Figure~\ref{fig:qualitative} further illustrate the typical failure modes of existing models, including under-segmentation and over-segmentation, while demonstrating our method's superior ability to accurately identify all event occurrences with precise temporal boundaries.

\textbf{Results on One-to-One Temporal Grounding.}
To verify that our approach does not compromise performance on conventional single-segment grounding, we evaluate on the TimeLens~\cite{timelens} benchmark (Table~\ref{tab:ootg_main}). Our OMTG-4B consistently improves upon both the base model and domain-specific baselines across all three datasets. 

Notably, the RL stage, trained exclusively on OMTG data without any one-to-one supervision, yields further gains over SFT across all benchmarks, which suggests that the one-to-many formulation cultivates more generalizable temporal grounding capabilities.

\subsection{Ablation Studies} 
\label{sec:ablation}

\textbf{Ablation on Training Strategy.} As shown in Table~\ref{tab:4b_performance_gain}, the Qwen3VL-4B base model demonstrates minimal capability on the OMTG task, achieving near-zero performance (EtF1: 0.21). However, SFT with our OMTG dataset fundamentally enables OMTG ability, significantly improving EtF1 to 34.81. Subsequently, the RL stage further improves this capability, boosting EtF1 to 43.65 (+8.84). Crucially, this progressive improvement confirms that while SFT establishes foundational OMTG ability, reinforcement learning with our proposed temporal and caption rewards provides additional alignment beyond supervised training alone.

\textbf{Ablation on RL Reward Design.} As shown in Table~\ref{tab:ablation_reward}, adding $R_{\text{tIoU}}$ yields consistent improvements in localization metrics. Adding $R_{\text{C-Acc}}$ further improves counting accuracy and EtF1, indicating that a direct cardinality signal benefits event perception. Incorporating $R_{\text{Caption}}$ achieves the best performance, improving C-Acc by 11.57 and EtF1 by 8.84 over the SFT baseline. We attribute this to dense captioning, which enforces fine-grained temporal perception and requires the model to explicitly distinguish each event occurrence. Note that $R_{\text{Caption}}$ is used in conjunction with $R_{\text{Length}}$, which prevents excessively verbose outputs that could dilute query-relevant temporal cues. Based on these findings, we adopt \textbf{tIoU + C-Acc + Caption + Length} as our default configuration.

\section{Conclusion}
\label{sec:conclusion}
In this paper, we identify and formalize the task of One-to-Many Temporal Grounding (OMTG), addressing the critical gap between current one-to-one paradigms and dynamic real-world scenarios. 

We reveal that existing state-of-the-art MLLMs, despite their success in standard settings, struggle significantly to perceive event cardinality and localize disjoint segments. 

To bridge this gap, we curate 56k high-quality one-to-many training samples via a sophisticated data pipeline and conduct SFT+RL training that incorporates our temporal and caption rewards, achieving state-of-the-art results. 

Our study establishes a strong baseline for this novel OMTG setting and facilitates future research in this direction.

\noindent
\textbf{Limitations and Future Works.}
Our current approach incurs high training costs and faces scalability challenges with extremely long videos. Future works will explore OMTG with memory in long videos setting.

\section*{Impact Statement}
This paper presents work aimed at advancing the field of Machine Learning, specifically in fine-grained video understanding and retrieval. Our proposed One-to-Many Temporal Grounding framework has the potential to positively impact society, such as enhancing video search efficiency, automating video editing workflows, and improving content accessibility. However, we acknowledge that advancements in precise temporal localization could potentially be misused in surveillance or privacy-intrusive applications. We explicitly condemn any use of our technology that violates individual privacy or human rights. We encourage the community to prioritize data privacy and responsible deployment when applying these technologies to sensitive real-world scenarios.

\bibliography{example_paper}
\bibliographystyle{icml2026}

\onecolumn
\appendix
\definecolor{headerblue}{RGB}{13, 110, 253}
\definecolor{lightblue}{RGB}{232, 240, 254}
\definecolor{lightyellow}{RGB}{255, 248, 220}
\definecolor{lightgreen}{RGB}{220, 240, 220}
\definecolor{lightgray}{RGB}{248, 249, 250}
\section*{Appendix Overview}
\begin{itemize}
    \item \textbf{\cref{sec:appendix_datapipeline}}: gives more details on training data pipeline building process.
   \item \textbf{\cref{sec:appendix_reward}}: presents more details on our reward function designs.
   \item \textbf{\cref{sec:videomme_results}}: shows the results on Video MME benchmark.
    \item \textbf{\cref{sec:modelsize_appendix}}: shows the results across different model size.
    \item \textbf{\cref{sec:implementation_details_benchmark_mllms}}: give the details of OMTG benchmarking process.
    \item \textbf{\cref{sec:ood}}: presents zero-shot OOD evaluation on longer in-the-wild videos.
    \item \textbf{\cref{sec:benchmark_statistics}}: analyzes the statistics details.
    \item \textbf{\cref{sec:failure_case}}: analyzes failure cases
    \item \textbf{\cref{sec:annotation_manual}}: shows the annotation process of OMTG and the human check process.
\end{itemize}

\newpage
\section{More Details of Training Data Pipeline}
\label{sec:appendix_datapipeline}
\subsection{Prompt Templates}
\label{sec:prompt}

In this section, we provide the detailed prompt templates for each stage in our proposed data pipeline.

\begin{table}[h]
\centering
\small
\setlength{\tabcolsep}{8pt}
\renewcommand{\arraystretch}{1.4}
\caption{Prompt template for Stage 1: Repetitive Event Discovery.}
\begin{tabular}{p{0.92\textwidth}}
\rowcolor{headerblue}
\textcolor{white}{\textbf{Stage 1: Repetitive Event Discovery}} \\[0.3em]
\rowcolor{lightgray}
The task is Repetitive Event Discovery. You need to scan the raw video to identify salient events that occur multiple times (repetitive events). \\[0.8em]

\rowcolor{lightblue}
\textbf{Task} \\
\rowcolor{lightblue}
\quad Based on the content of the video, generate descriptive queries for these repetitive events. \\
\rowcolor{lightblue}
\quad Place queries in \texttt{<query></query>} tags. \\
\rowcolor{lightblue}
\quad If no salient repetitive events exist, enter ``No'' in \texttt{<judge></judge>}; otherwise enter ``Yes''. \\[0.5em]

\rowcolor{lightgray}
\textbf{Requirements} \\[0.3em]
\rowcolor{lightyellow}
\textbf{(1) Repetition Requirement} \\
\rowcolor{lightyellow}
\quad The identified event must occur at least twice as distinct instances. \\
\rowcolor{lightyellow}
\quad It should NOT be a continuous state lasting the entire video. \\
\rowcolor{lightyellow}
\quad \textit{Example: Do not select ``a man standing'' if he stands there the whole time.} \\[0.5em]

\rowcolor{lightyellow}
\textbf{(2) Format Requirement} \\
\rowcolor{lightyellow}
\quad Output must be descriptive: phrase or short sentence. \\
\rowcolor{lightyellow}
\quad Format: ``subject + action (+ object/environment)'' \\
\rowcolor{lightyellow}
\quad \textit{Examples: ``a person jumping over a fence'', ``a dog catching a frisbee''} \\[0.5em]

\rowcolor{lightyellow}
\textbf{(3) Salience Requirement} \\
\rowcolor{lightyellow}
\quad Query should correspond to main content or significant actions. \\
\rowcolor{lightyellow}
\quad Must serve as meaningful entry for event localization. \\[0.5em]

\rowcolor{lightgreen}
\textbf{Output Format Examples} \\
\rowcolor{lightgreen}
\quad \textit{No repetitive events:} \texttt{<judge>No</judge>} \\
\rowcolor{lightgreen}
\quad \textit{With repetitive events:} \\
\rowcolor{lightgreen}
\quad \texttt{<judge>Yes</judge><query>a person jumping over a fence</query>} \\
\rowcolor{lightgreen}
\quad \texttt{<query>basketball player shooting a three-pointer</query>} \\
\bottomrule
\end{tabular}
\label{tab:stage1_prompt}
\end{table}

\begin{table}[h!]
\centering
\small
\setlength{\tabcolsep}{8pt}
\renewcommand{\arraystretch}{1.4}
\caption{Prompt template for Stage 2: Initial One-to-Many Grounding.}
\begin{tabular}{p{0.92\textwidth}}
\rowcolor{headerblue}
\textcolor{white}{\textbf{Stage 2: Initial One-to-Many Grounding}} \\[0.3em]
\rowcolor{lightgray}
Given a textual query, determine when the described content occurs in the video. \\[0.8em]

\rowcolor{lightblue}
\textbf{Input} \\
\rowcolor{lightblue}
\quad Textual Query: ``\{query\}'' \\[0.5em]

\rowcolor{lightgray}
\textbf{Task} \\
\rowcolor{lightgray}
\quad Identify all temporal segments where the queried event occurs. \\
\rowcolor{lightgray}
\quad Return timestamps in seconds. \\
\rowcolor{lightgray}
\quad If the specified query occurs multiple times, output multiple relevant time segments. \\[0.5em]

\rowcolor{lightgreen}
\textbf{Output Format} \\
\rowcolor{lightgreen}
\quad Return start and end timestamps for each occurrence. \\
\bottomrule
\end{tabular}
\label{tab:stage2_prompt}
\end{table}

\begin{table}[h!]
\centering
\small
\setlength{\tabcolsep}{8pt}
\renewcommand{\arraystretch}{1.2}
\caption{Prompt template for Stage 3: Strict Visual Verification.}
\begin{tabular}{p{0.92\textwidth}}
\rowcolor{headerblue}
\textcolor{white}{\textbf{Stage 3: Strict Visual Verification}} \\[0.3em]
\rowcolor{lightgray}
Verify whether the video segment satisfies the conditions described in the textual query. \\[0.8em]

\rowcolor{lightblue}
\textbf{Input} \\
\rowcolor{lightblue}
\quad Textual Query: ``\{query\}'' \\
\rowcolor{lightblue}
\quad Video Segment: [extracted segment from Stage 2] \\[0.5em]

\rowcolor{lightgray}
\textbf{Task} \\
\rowcolor{lightgray}
\quad Determine if the content in the video segment perfectly and completely \\
\rowcolor{lightgray}
\quad satisfies ALL conditions described in the textual query. \\[0.5em]

\rowcolor{lightyellow}
\textbf{Decision Criteria} \\
\rowcolor{lightyellow}
\quad Answer ``Yes'' if and only if ALL conditions are met. \\
\rowcolor{lightyellow}
\quad Answer ``No'' otherwise. \\[0.5em]

\rowcolor{lightgreen}
\textbf{Output Format} \\
\rowcolor{lightgreen}
\quad Binary response: ``Yes'' or ``No'' \\
\bottomrule
\end{tabular}
\label{tab:stage3_prompt}
\end{table}

\begin{table}[h!]
\centering
\small
\setlength{\tabcolsep}{8pt}
\renewcommand{\arraystretch}{1.2}
\caption{Prompt template for Stage 4: Recall Check and Query Refinement.}
\begin{tabular}{p{0.92\textwidth}}
\rowcolor{headerblue}
\textcolor{white}{\textbf{Stage 4: Recall Check and Query Refinement.}} \\[0.3em]
\rowcolor{lightgray}
You are an expert video temporal grounding annotator with exceptional attention to detail. Your task is to verify and refine temporal annotations for a specific query in a video. \\[0.8em]

\rowcolor{lightblue}
\textbf{Context} \\
\rowcolor{lightblue}
\quad Query: ``\{query\}'' \\
\rowcolor{lightblue}
\quad Dense Video Caption: \{dense\_caption\} \\
\rowcolor{lightblue}
\quad Previous Prediction: \{previous\_prediction\} \\[0.5em]

\rowcolor{lightgray}
\textbf{Task} \\
\rowcolor{lightgray}
\quad Carefully watch the video and identify ALL segments where the query occurs. \\[0.3em]
\rowcolor{lightgray}
\quad 1. \textbf{Verify existing predictions}: Check if previously annotated segments contain the queried event \\
\rowcolor{lightgray}
\quad 2. \textbf{Find missing segments}: Identify any additional occurrences that were missed \\
\rowcolor{lightgray}
\quad 3. \textbf{Refine boundaries}: Adjust start/end times to precisely capture event timing \\
\rowcolor{lightgray}
\quad 4. \textbf{Remove false positives}: Exclude segments that don't match the query \\
\rowcolor{lightgray}
\quad 5. \textbf{Refine query if needed}: If original query doesn't match but similar action occurs, refine it \\[0.5em]

\rowcolor{lightyellow}
\textbf{Critical Guidelines} \\
\rowcolor{lightyellow}
\quad $\bullet$ Watch the ENTIRE video carefully before making annotations \\
\rowcolor{lightyellow}
\quad $\bullet$ The query may appear multiple times --- find ALL occurrences \\
\rowcolor{lightyellow}
\quad $\bullet$ Be precise with timestamps --- round to the nearest second \\
\rowcolor{lightyellow}
\quad $\bullet$ Only include segments where the query is CLEARLY happening \\
\rowcolor{lightyellow}
\quad $\bullet$ Consider semantic equivalence (e.g., ``cleaning carpet'' includes scrubbing, vacuuming) \\
\rowcolor{lightyellow}
\quad $\bullet$ Do NOT include segments where someone talks about the action without performing it \\
\rowcolor{lightyellow}
\quad $\bullet$ Continuous actions should be ONE segment, not multiple 1-second segments \\
\rowcolor{lightyellow}
\quad $\bullet$ Segments should have meaningful duration (typically at least 2--3 seconds) \\[0.5em]

\rowcolor{lightyellow}
\textbf{Query Refinement Guidelines} \\
\rowcolor{lightyellow}
\quad $\bullet$ If exact query doesn't appear, check for SIMILAR action around predicted timestamps \\
\rowcolor{lightyellow}
\quad $\bullet$ Refined query should be concise and descriptive \\
\rowcolor{lightyellow}
\quad $\bullet$ \textit{Examples: ``person peeling egg'' $\rightarrow$ ``person breaking egg with hands''} \\
\rowcolor{lightyellow}
\quad $\bullet$ Only refine if action is genuinely similar/related \\[0.5em]

\rowcolor{lightgreen}
\textbf{Output Format} \\
\rowcolor{lightgreen}
\quad Respond with ONLY a JSON object: \\
\rowcolor{lightgreen}
\quad \texttt{\{``original\_query'': ``...'', ``query\_refined'': true/false,} \\
\rowcolor{lightgreen}
\quad \texttt{``refined\_query'': ``...'', ``segments'': [\{``start'': int, ``end'': int\}, ...],} \\
\rowcolor{lightgreen}
\quad \texttt{``reasoning'': ``...''\}} \\
\bottomrule
\end{tabular}
\label{tab:stage4_prompt}
\end{table}

\begin{table}[h!]
\centering
\small
\setlength{\tabcolsep}{8pt}
\renewcommand{\arraystretch}{1.4}
\caption{Prompt template for Stage 5: Query-Guided Dense Captioning.}
\begin{tabular}{p{0.92\textwidth}}
\rowcolor{headerblue}
\textcolor{white}{\textbf{Stage 5: Query-Guided Dense Captioning}} \\[0.3em]
\rowcolor{lightgray}
Analyze the given video and capture all distinct activity events occurring within it. For each event, provide a clear, descriptive label and specify its exact time interval. \\[0.8em]

\rowcolor{lightblue}
\textbf{Input} \\
\rowcolor{lightblue}
\quad Video: [input video] \\
\rowcolor{lightblue}
\quad Reference Queries: ``\{query\}'' \\[0.5em]

\rowcolor{lightgray}
\textbf{Important Requirements} \\[0.3em]
\rowcolor{lightyellow}
\textbf{(1) Fine-grained Timestamps} \\
\rowcolor{lightyellow}
\quad Break down long continuous activities into smaller, meaningful segments whenever possible. \\[0.3em]

\rowcolor{lightyellow}
\textbf{(2) Perceptible Changes} \\
\rowcolor{lightyellow}
\quad Each segment should represent a perceptible change in action, behavior, or context. \\[0.3em]

\rowcolor{lightyellow}
\textbf{(3) Temporal Continuity} \\
\rowcolor{lightyellow}
\quad Events should collectively cover the entire video without gaps or overlaps. \\[0.3em]

\rowcolor{lightyellow}
\textbf{(4) Query Integration} \\
\rowcolor{lightyellow}
\quad Include information from the given queries; describe these in detail within the content. \\
\rowcolor{lightyellow}
\quad \textit{Note: Do NOT directly copy queries --- they require more granular refinement.} \\[0.3em]

\rowcolor{lightyellow}
\textbf{(5) Precision over Brevity} \\
\rowcolor{lightyellow}
\quad Prioritize precision and semantic relevance over brevity. \\[0.3em]

\rowcolor{lightyellow}
\textbf{(6) Detailed Captions} \\
\rowcolor{lightyellow}
\quad Make generated captions as detailed as possible. \\
\rowcolor{lightyellow}
\quad Avoid overly broad or prolonged time intervals. \\
\rowcolor{lightyellow}
\quad \textit{Guideline: No single event should span more than 10--15 seconds unless clearly justified.} \\[0.5em]

\rowcolor{lightgreen}
\textbf{Output Format} \\
\rowcolor{lightgreen}
\quad For each event: descriptive label + time interval as ``start -- end seconds'' \\
\bottomrule
\end{tabular}
\label{tab:stage5_prompt}
\end{table}

\FloatBarrier
\newpage
\subsection{Quality Control}
\label{sec:quality_control}

To ensure the semantic consistency of our dataset, we implement a \textbf{Strict Visual Check} mechanism following the initial grounding (Stage 2). Given the query $Q$ and the set of predicted segments $\mathcal{S}=\{s_1, s_2, ..., s_N\}$ from Gemini 2.5 Pro, we employ a powerful open-source MLLM, Qwen3-VL-235B, as the verifier.

\textbf{Mechanism.} 
As outlined in Algorithm \ref{alg:visual_check}, the process operates on a "one-vote veto" principle. For a sample to be retained, \textit{every} individual segment $s_i$ must pass the visual verification against the query $Q$. If any segment is deemed irrelevant by the verifier, the entire sample is discarded.

\begin{algorithm}[h]
   \caption{Strict Visual Check Pipeline}
   \label{alg:visual_check}
\begin{algorithmic}
   \STATE {\bfseries Input:} Video $V$, Query $Q$, Segments $\mathcal{S}=\{s_1, ..., s_N\}$
   \STATE {\bfseries Model:} Verifier $\mathcal{M}$ (Qwen3-VL-235B)
   \STATE {\bfseries Output:} Boolean (Keep or Discard)
   \STATE
   \FOR{$i = 1$ {\bfseries to} $N$}
       \STATE $v_i \leftarrow \text{CropVideo}(V, s_i)$ \COMMENT{Extract video clip}
       \STATE $result \leftarrow \mathcal{M}(v_i, Q)$ \COMMENT{Verify alignment}
       \IF{$result$ is \text{Negative}}
           \STATE \textbf{return} \textbf{Discard} \COMMENT{One-vote veto}
       \ENDIF
   \ENDFOR
   \STATE \textbf{return} \textbf{Keep}
\end{algorithmic}
\end{algorithm}

\textbf{Theoretical Proof of Quality Gain}

We define the \textbf{Quality Gain} as the relative improvement of the sample validity probability after passing the visual check compared to the raw probability.

\noindent\textbf{Formulation.}
Let $\theta$ be the prior probability of a segment mismatch, and $p$ be the verifier's error rate. Based on the independence assumption:
\begin{itemize}
    \item \textbf{Prior Validity (No Check):} $P(\text{Valid}) = (1-\theta)^N$
    \item \textbf{Posterior Validity (Passed Check):} $P(\text{Valid} | \text{Pass}) = \left( \frac{(1-\theta)(1-p)}{(1-\theta)(1-p) + \theta p} \right)^N$
\end{itemize}

\noindent\textbf{Quantifying the Improvement.}
We measure the improvement using the \textbf{Relative Lift} Quality Gain ($\mathcal{L}$), defined as the ratio of the posterior to the prior:
\begin{equation}
    \mathcal{L}(N) = \frac{P(\text{Valid} | \text{Pass})}{P(\text{Valid})} = \left( \frac{1-p}{1-\theta-p} \right)^N
\end{equation}
Let the base term be $\beta = \frac{1-p}{1-\theta-p}$. As proved previously, if $0<p<1$ and $0<\theta<1$ and $0<1-\theta-p<1$, then $\beta > 1$. Consequently, the lift $\mathcal{L}(N) = \beta^N$ grows exponentially with $N$. This implies that the visual check is significantly more effective at filtering noise for complex samples (higher $N$) than for simple ones.

\noindent\textbf{Numerical Analysis ($N=2$ vs. $N=4$).}
Based on our statistics, let $\theta \approx 0.5$ (raw data noise, we roughly assume this equals to $1 - \text{C-Acc}$ of Gemini 2.5 pro) and $p \approx 0.2$ (verifier error).
\begin{itemize}
    \item \textbf{Base Term $\beta$:}
    \begin{equation}
        \beta = \frac{1-0.2}{1-0.5-0.2} = 2.67
    \end{equation}
    \item \textbf{Improvement for $N=2$:}
    \begin{equation}
        \mathcal{L}(2) = 2.67^2 \approx \mathbf{7.13\times} \quad (\text{Quality boosted by 7 times})
    \end{equation}
    \item \textbf{Improvement for $N=4$:}
    \begin{equation}
        \mathcal{L}(4) = 2.67^4 \approx \mathbf{50.82\times} \quad (\text{Quality boosted by 50 times})
    \end{equation}
\end{itemize}
The calculation demonstrates that the quality improvement for $N=4$ is substantially higher than for $N=2$.

Given this and based on our experimental results, we directly accept samples with $N \ge 4$ without further processing. Conversely, for samples with fewer segments ($N=2, 3$), we employ \textit{Stage 4: Recall Check and Query Refinement} to further boost data quality. Experiments in Tab.~\ref{tab:appendix_data_ablatoin} have shown that this strategy is very effective in improving data quality.

\begin{table}[h!]
\centering
\small
\setlength{\tabcolsep}{2.8pt}  
\renewcommand{\arraystretch}{1.0} 
\caption{\textbf{Performance comparison between using data w/ or w/o quality control.}}
\begin{tabular}{lcccccc}
\toprule
\textbf{Model} & \textbf{C-Acc} & \textbf{tF1@0.3} & \textbf{tF1@0.5} & \textbf{tF1@0.7} & \textbf{tIoU} & \textbf{EtF1} \\ \midrule
SFT data w/o quality control  & 39.17 & 66.32 & 59.43 & 40.65 & 53.82 & 29.73 \\
RL data w/o quality control  & 40.25 & 65.41 & 60.01 & 41.72 & 54.14 & 32.55 \\
\midrule
SFT data w/ quality control  & 44.06 & 69.57 & 61.23 & 45.63 & 56.94 & 34.81 \\
RL data w/ quality control  & \textbf{55.63} & \textbf{73.46} & \textbf{65.40} & \textbf{48.96} & \textbf{61.24} & \textbf{43.65} \\
\bottomrule
\end{tabular}
\label{tab:appendix_data_ablatoin}
\end{table}

\section{More Details of Reward Functions Design}
\label{sec:appendix_reward}

In this section, we provide the detailed prompt templates for caption reward evaluation and the formulation of the length penalty. 

\subsection{Caption Reward Prompts}
\label{sec:caption_reward}
The caption reward $R_{\text{Caption}}$ employs a two-part LLM-as-Judge evaluation framework using Qwen3-30B-A3B as the reward model. Both parts are computed in parallel to assess complementary aspects of caption quality.

\noindent\textbf{Part 1: Caption Quality Score.}
The Caption Quality Score ($S_{\text{cq}}$) evaluates captions with access to ground truth annotations across three dimensions. \textit{Coverage} ($S_{\text{cov}}$) measures what fraction of ground truth segments are matched by corresponding captions with appropriate descriptions. \textit{Precision} ($S_{\text{prec}}$) assesses how accurately the temporal boundaries of captions align with ground truth intervals, penalizing both undershooting and overshooting. \textit{Discriminability} ($S_{\text{disc}}$) determines whether each caption provides unique contextual information (e.g., who, what, when, where) to distinguish different occurrences of the same event. The composite score is computed as:
\begin{equation}
S_{\text{cq}} = \mu_1 \cdot S_{\text{cov}} + \mu_2 \cdot S_{\text{prec}} + \mu_3 \cdot S_{\text{disc}}
\end{equation}
where $\mu_1=0.5$, $\mu_2=0.3$, and $\mu_3=0.2$ to emphasize coverage completeness. The prompt template is shown in Table~\ref{tab:caption_quality_prompt}.

\noindent\textbf{Part 2: Caption Guided Grounding Score.}
The Caption Guided Grounding Score ($S_{\text{cgg}}$) evaluates whether the generated captions contain sufficient information for event localization. Given only the text query and generated captions (without access to the video), the judge identifies all segments where the queried event likely occurs by matching caption descriptions to the query semantics. The predicted intervals are then compared against ground truth using tF1 scores at IoU thresholds of 0.3 and 0.5:
\begin{equation}
S_{\text{cgg}} = \frac{\text{tF1}@0.3 + \text{tF1}@0.5}{2}
\end{equation}
This text-only grounding evaluation ensures that captions are semantically informative rather than merely temporally co-occurring with ground truth. The prompt template is shown in Table~\ref{tab:caption_grounding_prompt}.

The final caption reward combines the two components:
\begin{equation}
    R_{\text{Caption}} = \alpha \cdot S_{\text{cq}} + (1-\alpha) \cdot S_{\text{cgg}}
\end{equation}
where $\alpha=0.5$ balances quality assessment and grounding consistency.

\subsection{Length Penalty}
\label{sec:length_penalty}

Excessively long responses can introduce irrelevant details, dilute query-relevant temporal cues, and degrade localization performance. We adopt a soft length penalty that progressively penalizes responses exceeding predefined thresholds.

\vspace{0.5em}
\noindent\textbf{Soft Overlong Punishment.}
For a given text length $L$, we define the soft overlong penalty function as:
\begin{equation}
P(L; L_{\text{soft}}, L_{\text{hard}}, \alpha) = 
\begin{cases}
0 & \text{if } L \leq L_{\text{soft}} \\[6pt]
\alpha \cdot \dfrac{L - L_{\text{soft}}}{L_{\text{hard}} - L_{\text{soft}}} & \text{if } L_{\text{soft}} < L \leq L_{\text{hard}} \\[6pt]
\alpha & \text{if } L > L_{\text{hard}}
\end{cases}
\end{equation}
where $L_{\text{soft}}$ is the soft threshold below which no penalty is applied, $L_{\text{hard}}$ is the hard threshold beyond which the maximum penalty is reached, and $\alpha$ is the penalty factor.

\vspace{0.5em}
\noindent\textbf{Total Length Penalty.}
The total length penalty consists of two components.

\textit{Thinking Content Penalty} penalizes overly verbose reasoning in the \texttt{<think>} block:
\begin{equation}
P_{\text{think}} = P(L_{\text{think}}; 2000, 5000, 1.0)
\end{equation}
where $L_{\text{think}}$ is the character count of the thinking content.

\textit{Caption Length Penalty} is applied for excessively long captions. The average caption length penalty across all $N$ captions is:
\begin{equation}
P_{\text{cap}} = \frac{1}{N} \sum_{i=1}^{N} P(L_{\text{cap}}^{(i)}; 100, 200, 0.5)
\end{equation}
where $L_{\text{cap}}^{(i)}$ is the character count of the $i$-th caption.

The final length penalty is computed as:
\begin{equation}
R_{\text{Length}} = P_{\text{think}} + P_{\text{cap}}
\end{equation}

\definecolor{headerblue}{RGB}{13, 110, 253}
\definecolor{lightblue}{RGB}{232, 240, 254}
\definecolor{lightyellow}{RGB}{255, 248, 220}
\definecolor{lightgreen}{RGB}{220, 240, 220}
\definecolor{lightgray}{RGB}{248, 249, 250}

\begin{table}[h!]
\centering
\small
\setlength{\tabcolsep}{8pt}
\renewcommand{\arraystretch}{1.4}
\caption{Prompt template for $S_{\text{cq}}$ evaluation.}
\begin{tabular}{p{0.92\textwidth}}
\rowcolor{headerblue}
\textcolor{white}{\textbf{Part 1: $S_{\text{cq}}$ Evaluation Prompt}} \\[0.3em]
\rowcolor{lightgray}
You are a STRICT evaluator for Video Temporal Grounding caption quality. \\[0.8em]

\rowcolor{lightblue}
\textbf{Context} \\
\rowcolor{lightblue}
\quad Query: ``\{query\}'' \\
\rowcolor{lightblue}
\quad Ground Truth: \{num\_gt\_intervals\} segment(s) at \{gt\_intervals\_str\} \\
\rowcolor{lightblue}
\quad Video duration: approximately \{video\_duration\}s \\[0.5em]

\rowcolor{lightblue}
\textbf{Model's Captions} \\
\rowcolor{lightblue}
\quad \{caption\_list\_str\} \\[0.5em]

\rowcolor{lightgray}
\textbf{Evaluation Task} \\[0.3em]
\rowcolor{lightgray}
\textit{Step 1: Map each GT to captions} \\
\rowcolor{lightgray}
\quad For each GT segment, find the BEST matching caption (if any). \\
\rowcolor{lightgray}
\quad A match requires: (1) temporal overlap, AND (2) caption describes ``\{query\}'' \\[0.5em]
\rowcolor{lightgray}
\textit{Step 2: Score STRICTLY using these rules} \\[0.5em]

\rowcolor{lightyellow}
\textbf{$S_{\text{cov}}$ (0--10)}: What fraction of GT segments are matched? \\
\rowcolor{lightyellow}
\quad 10 = ALL \{num\_gt\_intervals\} GT matched with clear ``\{query\}'' descriptions \\
\rowcolor{lightyellow}
\quad 8 = ALL matched, but one has weak description \\
\rowcolor{lightyellow}
\quad 6 = approx.\ 70\% matched \quad\quad 4 = approx.\ 50\% matched \\
\rowcolor{lightyellow}
\quad 2 = Only one matched \quad\quad 0 = None matched \\
\rowcolor{lightyellow}
\quad \textit{Note: If ANY GT is missing, score at most 8} \\[0.5em]

\rowcolor{lightyellow}
\textbf{$S_{\text{prec}}$ (0--10)}: How close are boundaries? \\
\rowcolor{lightyellow}
\quad 10 = ALL within 1s of GT \quad 8 = Most within 2s \quad 6 = Within 3--5s \\
\rowcolor{lightyellow}
\quad 4 = Off by 5--10s \quad 2 = Off by more than 10s \\
\rowcolor{lightyellow}
\quad \textit{Note: Captions much WIDER than GT count as imprecise} \\[0.5em]

\rowcolor{lightyellow}
\textbf{$S_{\text{disc}}$ (0--10)}: Can occurrences be distinguished? \\
\rowcolor{lightyellow}
\quad 10 = Each has a unique context (who/what/when/where) \\
\rowcolor{lightyellow}
\quad 7 = Good context for most \quad 4 = Generic \quad 0 = Impossible to distinguish \\[0.5em]

\rowcolor{lightgreen}
\textbf{Output Format} \\
\rowcolor{lightgreen}
\quad After analysis, output ONLY valid JSON: \\
\rowcolor{lightgreen}
\quad \texttt{\{``coverage'': int 0--10, ``precision'': int 0--10, ``discriminability'': int 0--10\}} \\
\rowcolor{lightgreen}
\quad BE STRICT: Average captions score 4--6, not 8--10. \\
\bottomrule
\end{tabular}
\label{tab:caption_quality_prompt}
\end{table}

\begin{table}[h!]
\centering
\small
\setlength{\tabcolsep}{8pt}
\renewcommand{\arraystretch}{1.4}
\caption{Prompt template for $S_{\text{cgg}}$ evaluation.}
\begin{tabular}{p{0.92\textwidth}}
\rowcolor{headerblue}
\textcolor{white}{\textbf{Part 2: $S_{\text{cgg}}$ Evaluation Prompt}} \\[0.3em]
\rowcolor{lightgray}
You are predicting video timestamps from text captions ONLY (no video access). \\[0.8em]

\rowcolor{lightblue}
\textbf{Query}: ``\{query\}'' \\[0.5em]

\rowcolor{lightblue}
\textbf{Captions} \\
\rowcolor{lightblue}
\quad \{caption\_list\_str\} \\[0.5em]

\rowcolor{lightgray}
\textbf{Task} \\
\rowcolor{lightgray}
\quad Find ALL segments where ``\{query\}'' occurs based on the captions. \\[0.5em]

\rowcolor{lightgray}
\textbf{Rules} \\
\rowcolor{lightgray}
\quad 1. Look for captions that DESCRIBE or IMPLY ``\{query\}'' \\
\rowcolor{lightgray}
\quad 2. Use the caption's timestamp as your prediction \\
\rowcolor{lightgray}
\quad 3. If multiple captions match, list all of them \\
\rowcolor{lightgray}
\quad 4. If caption text is vague but likely refers to the query, include it \\
\rowcolor{lightgray}
\quad 5. Output format: one segment per line as ``start -- end'' \\[0.5em]

\rowcolor{lightgreen}
\textbf{Example Output} \\
\rowcolor{lightgreen}
\quad \texttt{10.5 -- 15.0} \\
\rowcolor{lightgreen}
\quad \texttt{32.0 -- 37.0} \\[0.3em]
\rowcolor{lightgreen}
\quad Your predictions (list ALL matching segments): \\
\bottomrule
\end{tabular}
\label{tab:caption_grounding_prompt}
\end{table}

\subsection{Temporal Rewards Design Choices}
To identify the optimal supervision for the temporal branch, we conduct an ablation study on different combinations of temporal rewards: $\mathcal{R}_{\text{tIoU}}$, $\mathcal{R}_{\text{tF1}}$, and $\mathcal{R}_{\text{C-Acc}}$. The definitions of these rewards strictly follow the metrics defined in Section~\ref{sec:metric}. We report the performance gains over the SFT baseline in Tab~\ref{tab:ablation_reward_appendix}.

\begin{table}[ht]
\centering
\small
\setlength{\tabcolsep}{2.5pt} 
\renewcommand{\arraystretch}{1.1} 
\caption{\textbf{Ablation on different temporal reward combinations on OMTG Bench.} All results are reported as absolute improvements over the SFT baseline (Row 1). $\mathcal{R}_{\text{tIoU}} + \mathcal{R}_{\text{C-Acc}}$ yields the best balance between localization and cardinality.}
\begin{tabular}{lcccccc}
\toprule
\textbf{Reward Functions} & \textbf{C-Acc} & \textbf{tF1@30} & \textbf{tF1@50} & \textbf{tF1@70} & \textbf{tIoU} & \textbf{EtF1} \\ \midrule
$\mathcal{R}_{\text{tIoU}}$ & +0.31 & +2.01 & +1.82 & +0.28 & +2.61 & +0.74 \\ 
$\mathcal{R}_{\text{tIoU}} + \mathcal{R}_{\text{tF1}}$ & +1.88 & +3.65 & +3.15 & +1.71 & +3.12 & +2.96 \\
\rowcolor[HTML]{E8F0FB} 
$\mathcal{R}_{\text{tIoU}} + \mathcal{R}_{\text{C-Acc}}$ & \textbf{+9.06} & \textbf{+5.55} & \textbf{+4.86} & \textbf{+3.64} & \textbf{+5.23} & \textbf{+7.91} \\ 
$\mathcal{R}_{\text{tIoU}} + \mathcal{R}_{\text{tF1}} + \mathcal{R}_{\text{C-Acc}}$ & +5.94 & +2.94 & +3.62 & +3.29 & +3.21 & +5.50 \\
\bottomrule
\end{tabular}

\label{tab:ablation_reward_appendix}
\end{table}

\noindent\textbf{Critical Role of Cardinality Supervision.} 
As shown in Table~\ref{tab:ablation_reward}, using only the boundary-aware reward ($\mathcal{R}_{\text{tIoU}}$) yields negligible improvement in Count Accuracy (+0.31 C-Acc). This indicates that standard overlap-based objectives encourage the model to refine local boundaries but fail to rectify the number of predicted segments (e.g., merging two distinct events or splitting one event). However, incorporating the cardinality-aware reward ($\mathcal{R}_{\text{tIoU}} + \mathcal{R}_{\text{C-Acc}}$) results in a substantial performance leap, particularly in C-Acc (+9.06) and the comprehensive metric EtF1 (+7.91). This confirms that explicit supervision on event counts is indispensable for One-to-Many Temporal Grounding, as it forces the model to discern the discrete nature of multiple occurrences.

\noindent\textbf{Redundancy in Dense Temporal Rewards.} 
We further investigate the effect of adding the Temporal F1 reward ($\mathcal{R}_{\text{tF1}}$). Surprisingly, the combination of all three rewards ($\mathcal{R}_{\text{tIoU}} + \mathcal{R}_{\text{tF1}} + \mathcal{R}_{\text{C-Acc}}$) leads to a performance degradation compared to the simpler $\mathcal{R}_{\text{tIoU}} + \mathcal{R}_{\text{C-Acc}}$ setting. We hypothesize that $\mathcal{R}_{\text{tIoU}}$ and $\mathcal{R}_{\text{tF1}}$ provide overlapping supervision signals regarding localization quality. Optimizing these redundant, dense objectives simultaneously may dilute the gradient signal from the sparse cardinality reward $\mathcal{R}_{\text{C-Acc}}$. Consequently, we adopt $\mathcal{R}_{\text{tIoU}} + \mathcal{R}_{\text{C-Acc}}$ as our final temporal reward design choice.

\section{More Results on Video MME}
\label{sec:videomme_results}

To assess whether our specialized training for One-to-Many Temporal Grounding compromises the model's general video understanding capabilities, we evaluated our models on the \textbf{VideoMME}~\cite{fu2025videomme} benchmark. We report results under the setting without subtitles (w/o sub), sampling 128 frames per video. The results are summarized in Tab~\ref{tab:videomme}.

\begin{table}[ht]
\centering
\small
\setlength{\tabcolsep}{4pt}
\caption{\textbf{Results on VideoMME (w/o sub, 128 frames).} We compare our OMTG-4B variants against the backbone Qwen3-VL-4B to analyze the impact of our training strategies on general video understanding.}
\begin{tabular}{lcccc}
\toprule
\textbf{Model} & \textbf{Overall} & \textbf{Short} & \textbf{Medium} & \textbf{Long} \\ \midrule
Qwen3-VL-4B (Base) & \textbf{66.7} & \textbf{77.6} & \textbf{65.8} & \textbf{56.8} \\ \midrule
\textit{Ablation on SFT Training Data} & & & & \\
OMTG-4B (SFT w/o CoT) & 62.1 & 72.4 & 62.0 & 52.0 \\
OMTG-4B (SFT w/ CoT) & 64.4 & \textbf{77.6} & 64.0 & 51.8 \\ \midrule
\textit{Ablation on Reward Functions} & & & & \\
OMTG-4B (RL w/o Cap. Reward) & 62.5 & 72.7 & 62.3 & 52.4 \\
\textbf{OMTG-4B (RL w/ Cap. Reward)} & \underline{65.1} & 77.3 & \underline{65.7} & \underline{52.3} \\ \bottomrule
\end{tabular}
\label{tab:videomme}
\end{table}

As expected, domain-specific fine-tuning typically incurs a trade-off in general capabilities. The naive SFT model (w/o CoT) exhibits a performance drop compared to the backbone Qwen3-VL (62.1 vs. 66.7). However, our proposed strategies effectively mitigate this issue:

\textbf{Impact of CoT:} Incorporating Chain-of-Thought (CoT) data during SFT significantly recovers general performance (+2.3\% Overall), particularly in Short videos where it matches the backbone (77.6). This suggests that enhancing reasoning capabilities benefits both temporal grounding and general video understanding.

\textbf{Impact of RL with Caption Reward:} The RL stage further improves the performance. Notably, including the Caption Reward is crucial; it boosts the Overall score to \textbf{65.1}, narrowing the gap with the backbone to a minimal margin. This indicates that the Caption Reward helps the model maintain high-quality semantic representations while optimizing for grounding metrics.

In summary, our final OMTG-4B model evolves into a specialist in temporal grounding while remaining a robust generalist in video understanding.

\section{Performances across Different Model Sizes}
\label{sec:modelsize_appendix}
In this section, we present a analysis of how model capacity affects performance on OMTG Bench. As illustrated in Tab.~\ref{tab:ablation_modelsize_appendix}.

\begin{table}[h!]
\centering
\small
\setlength{\tabcolsep}{2.5pt} 
\renewcommand{\arraystretch}{1.} 
\caption{\textbf{Performances across Different Model Sizes}} 
\begin{tabular}{lcccccc}
\toprule
\textbf{Model Size} & \textbf{C-Acc} & \textbf{tF1@30} & \textbf{tF1@50} & \textbf{tF1@70} & \textbf{tIoU} & \textbf{EtF1} \\ \midrule
OMTG-2B & 46.56 & 68.32 & 61.06 & 44.10 & 55.97 & 35.48 \\ 

OMTG-4B & 55.63 & 73.46 & 65.40 & 48.96 & 61.24 & 43.65 \\
\bottomrule
\end{tabular}
\label{tab:ablation_modelsize_appendix}
\end{table}

\section{Implementation Details for OMTG Benchmarking}
\label{sec:implementation_details_benchmark_mllms}

In this section, we present the implementation details for evaluating existing MLLMs on our OMTG evaluation suite, yielding the results reported in Tab. \ref{tab:omtg_main}.

\textbf{Proprietary Models.} 
We evaluated the \textbf{Gemini series} (Gemini 2.5 Pro and Gemini 3 Pro) via their official Video Understanding API. Notably, the inputs for these models incorporated both visual and audio modalities to maximize information intake. 
For \textbf{Seed-1.8}, we accessed the model via the Volcano Engine API. We uploaded the complete video files and used the default sampling configuration, extracting frames at 2.0 FPS.

\textbf{Open-Source Models.} 
For our \textbf{OMTG-4B} model and the \textbf{Qwen series} (including Qwen3-VL and Qwen2.5-VL), we employed the \texttt{sglang} engine as the inference backend to ensure efficiency. 
For other open-source baselines, we utilized the standard \texttt{transformers} library for inference. 
To ensure a fair comparison regarding visual information, we imposed consistent resolution constraints across all open-source models (including ours), setting \texttt{fps=2}, \texttt{min\_pixels=2048}, and \texttt{total\_pixels=8388608}. For UniTime, we follow it's default adaptive frame scaling strategy setting.

\textbf{Prompts.} 
To ensure reproducibility, we strictly standardized the prompts used for evaluation:
    \begin{itemize}
        \item All open-source models (including our OMTG-4B) and Seed-1.8 utilized the unified prompt detailed in the Tab~\ref{tab:prompt_opensource}.
        \item The Gemini series utilized the specific prompt detailed in the Tab~\ref{tab:prompt_gemini}.
    \end{itemize}

\begin{table}[ht]
\centering
\small
\setlength{\tabcolsep}{8pt}
\renewcommand{\arraystretch}{1.4}
\caption{Prompt template for open-source models and Seed-1.8 evaluation.}
\begin{tabular}{p{0.92\textwidth}}
\rowcolor{headerblue}
\textcolor{white}{\textbf{Open-Source Models and Seed-1.8 Evaluation Prompt}} \\[0.3em]
\rowcolor{lightgray}
Find the video segment that corresponds to the given textual query '\{query\}' and determine its start and end seconds. If there are multiple segments,
please output the start and end time for each one separately. \\[0.8em]

\end{tabular}
\label{tab:prompt_opensource}
\end{table}

\begin{table}[ht]
\centering
\small
\setlength{\tabcolsep}{8pt}
\renewcommand{\arraystretch}{1.4}
\caption{Prompt template for Gemini series evaluation.}
\begin{tabular}{p{0.92\textwidth}}
\rowcolor{headerblue}
\textcolor{white}{\textbf{Gemini Series Evaluation Prompt}} \\[0.3em]
\rowcolor{lightgray}
Find the video segment that corresponds to the given textual query '\{query\}' and determine its start and end seconds. Format your response as: `<time>{{start}} - {{end}} seconds</time>`

Where:

- {{start}} = starting second

- {{end}} = ending second

Example: `<time>40 - 49 seconds</time>`. If there exists multiple segments, separate them with a comma, e.g., `<time>10 - 13 seconds</time>, <time>27 - 29 seconds</time>`. \\[0.8em]

\end{tabular}
\label{tab:prompt_gemini}
\end{table}

\section{In-the-Wild Generalization}
\label{sec:ood}

To further validate the generalization capability beyond the benchmark domain, we collect an additional out-of-domain (OOD) test set from recent videos on Bilibili and YouTube. Using the same data pipeline and human verification protocol as the main benchmark, we annotate 60 samples across 52 videos, covering diverse real-world content including travel vlogs, gaming, sports, news, and anime. 

These videos are entirely out-of-domain from any training source. The average duration is 422.87s (max 1419.93s). Despite the limited sample size, this set serves as a challenging OOD test for zero-shot evaluation.

Table~\ref{tab:ood} reports the zero-shot performance. Our model demonstrates strong generalization to longer, truly in-the-wild videos, significantly outperforming baselines across all metrics.

\begin{table}[h]
\centering
\small
\caption{Zero-shot OOD evaluation on in-the-wild videos.}
\label{tab:ood}
\begin{tabular}{lcccccccc}
\toprule
Model & tIoU & C-Acc & F1@0.3 & F1@0.5 & F1@0.7 & EtF1 \\
\midrule
Gemini-2.5-Pro & 12.99 & 15.00 & 16.35 & 12.17 & 7.15 & 4.81  \\
Gemini-3-Pro & 14.22 & 21.67 & 16.60 & 12.68 & 6.53 & 3.29 \\
Qwen3-VL-4B & 18.91 & 0.13 & 22.11 & 15.72 & 11.09 & 0.09 \\
\midrule
OMTG-4B (Ours) & \textbf{40.51} & \textbf{35.00} & \textbf{50.39} & \textbf{44.21} & \textbf{30.55} & \textbf{22.10} \\
\bottomrule
\end{tabular}
\end{table}

\section{Statistics Details of OMTG Benchmark}
\label{sec:benchmark_statistics}

In this section, we present more statistics of our OMTG Benchmark.

\begin{figure}[h!]
    \centering
    \begin{subfigure}[b]{0.4\textwidth}
        \centering
        \includegraphics[width=0.6\linewidth]{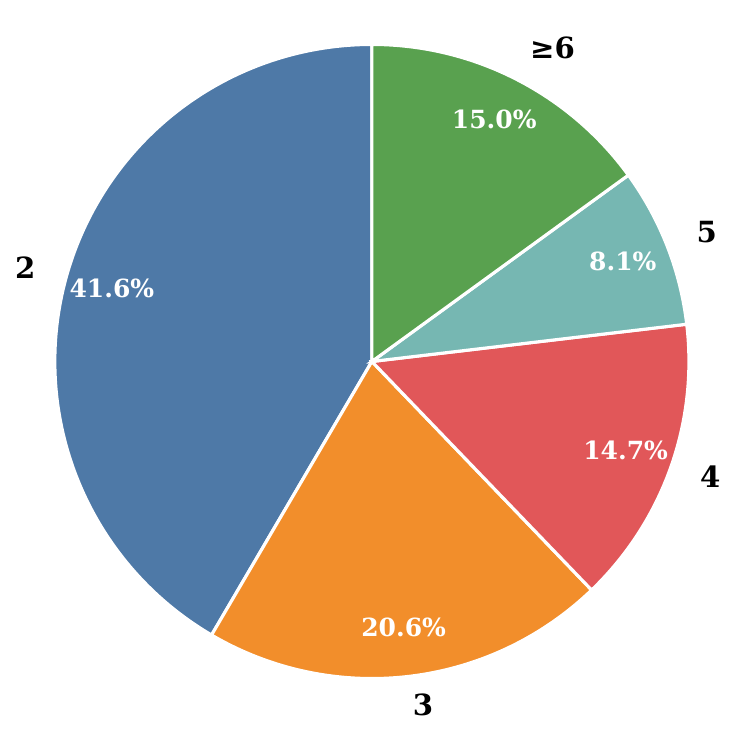}
        \caption{Distribution of \#ground truth segments.}
        \label{fig:segment_dist}
    \end{subfigure}
    \hfill 
    \begin{subfigure}[b]{0.4\textwidth}
        \centering
        \includegraphics[width=\linewidth]{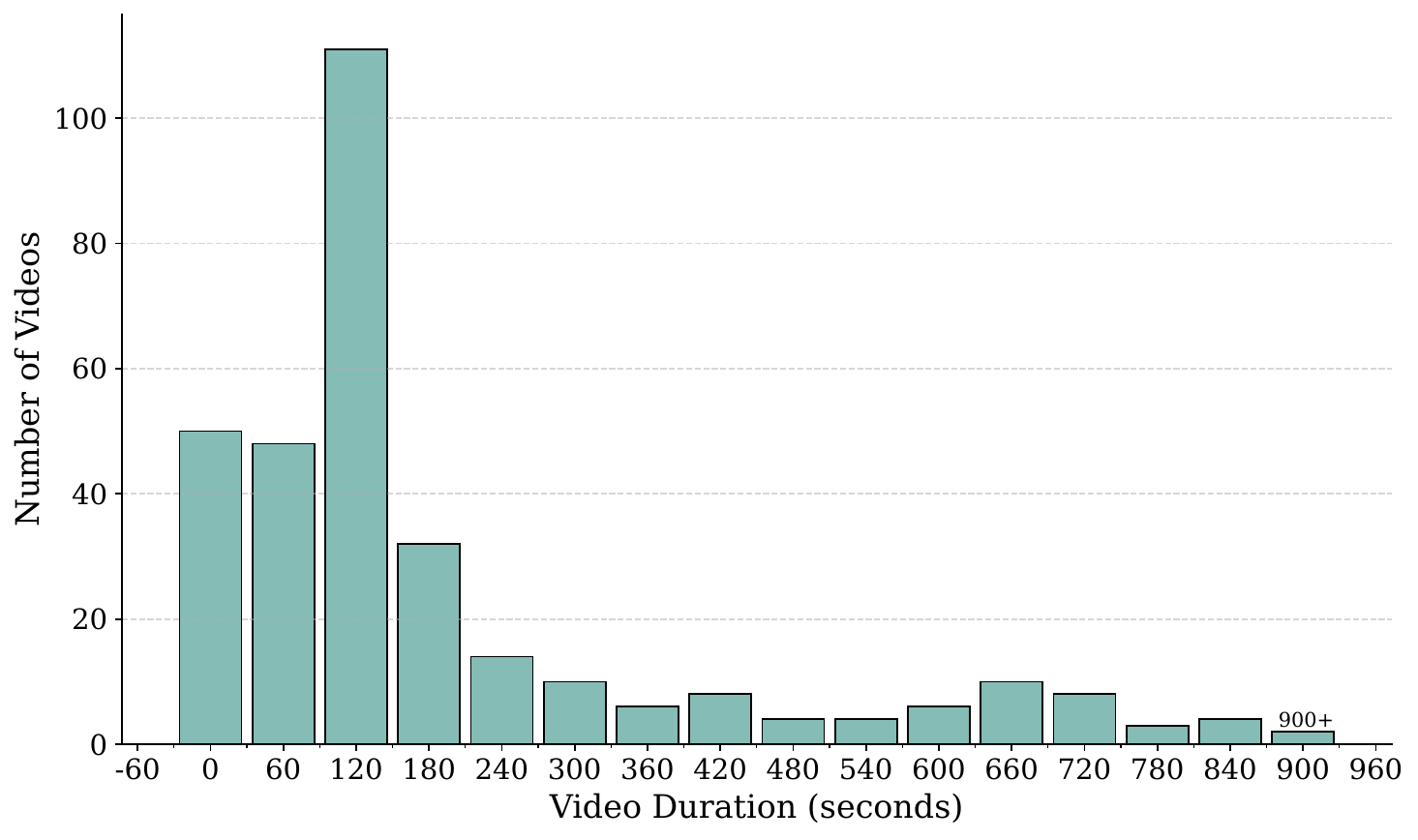}
        \caption{Distribution of video durations (in seconds).}
        \label{fig:duration_dist}
    \end{subfigure}
    \caption{\textbf{OMTG Benchmark Statistics.} Left: The distribution of the number of temporal segments in the ground truth. Right: The histogram of video durations in the benchmark.}
    \label{fig:bench_stats}
\end{figure}

\begin{figure}[h!]
    \centering
    \includegraphics[width=\textwidth]{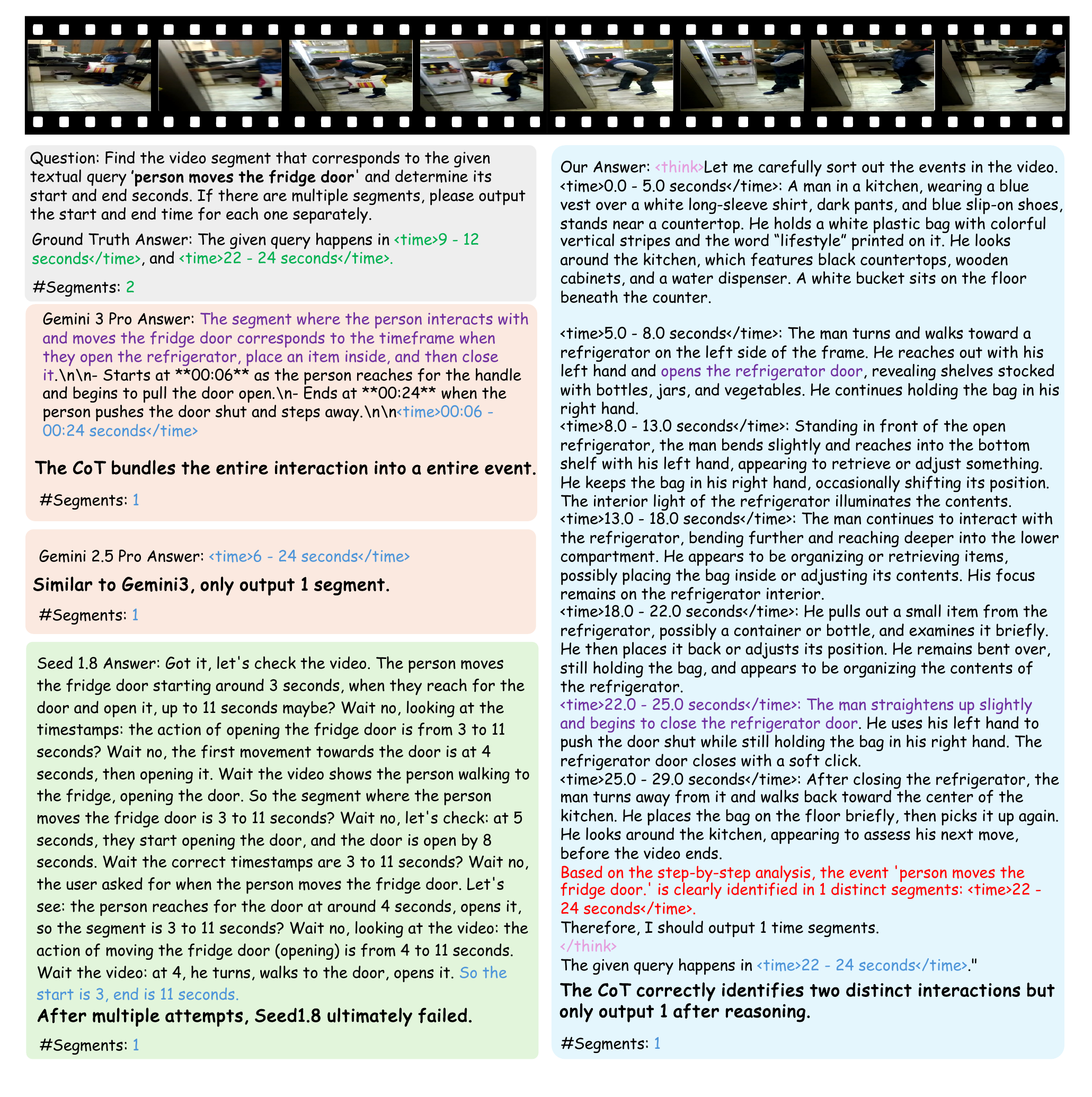}
    \caption{\textbf{Failure case from video T42IZ.mp4}}
    \label{fig:failure_case}
\end{figure}

\section{Failure Cases Study}
\label{sec:failure_case}
In this section, we present typical failure cases and provide analyses.

On hard cases like Fig. ~\ref{fig:failure_case} ("person moves the fridge door"), the model exhibits complete temporal misalignment when action semantics are ambiguous or visually subtle. The query "moves the fridge door" creates interpretive ambiguity—models struggle to distinguish between the transitional motion of opening/closing (discrete actions) and the sustained state of the door being open (static condition), often defaulting to coarse-grained segmentation (e.g., 06-24s encompassing the entire interaction) or fixating on visually salient but semantically irrelevant frames (e.g., the person walking toward the fridge rather than the hand manipulating the handle). This reveals a critical vulnerability: when action boundaries lack sharp visual distinctive features, the model prioritizes scene context over fine-grained motion semantics, leading to predictions that either dilute the precise temporal boundaries or completely drift away from the actual motion event.

\section{Annotation Interface and Manual Check For OMTG Benchmark}
\label{sec:annotation_manual}

To construct a high-quality One-to-Many Temporal Grounding (OMTG) benchmark, we developed a custom web-based annotation tool designed to facilitate the precise labeling of disjoint temporal segments. This section details the interface design, the annotation workflow, and the strict quality criteria provided to the annotators.

\begin{figure}[htbp]
    \centering
    \includegraphics[width=0.95\textwidth]{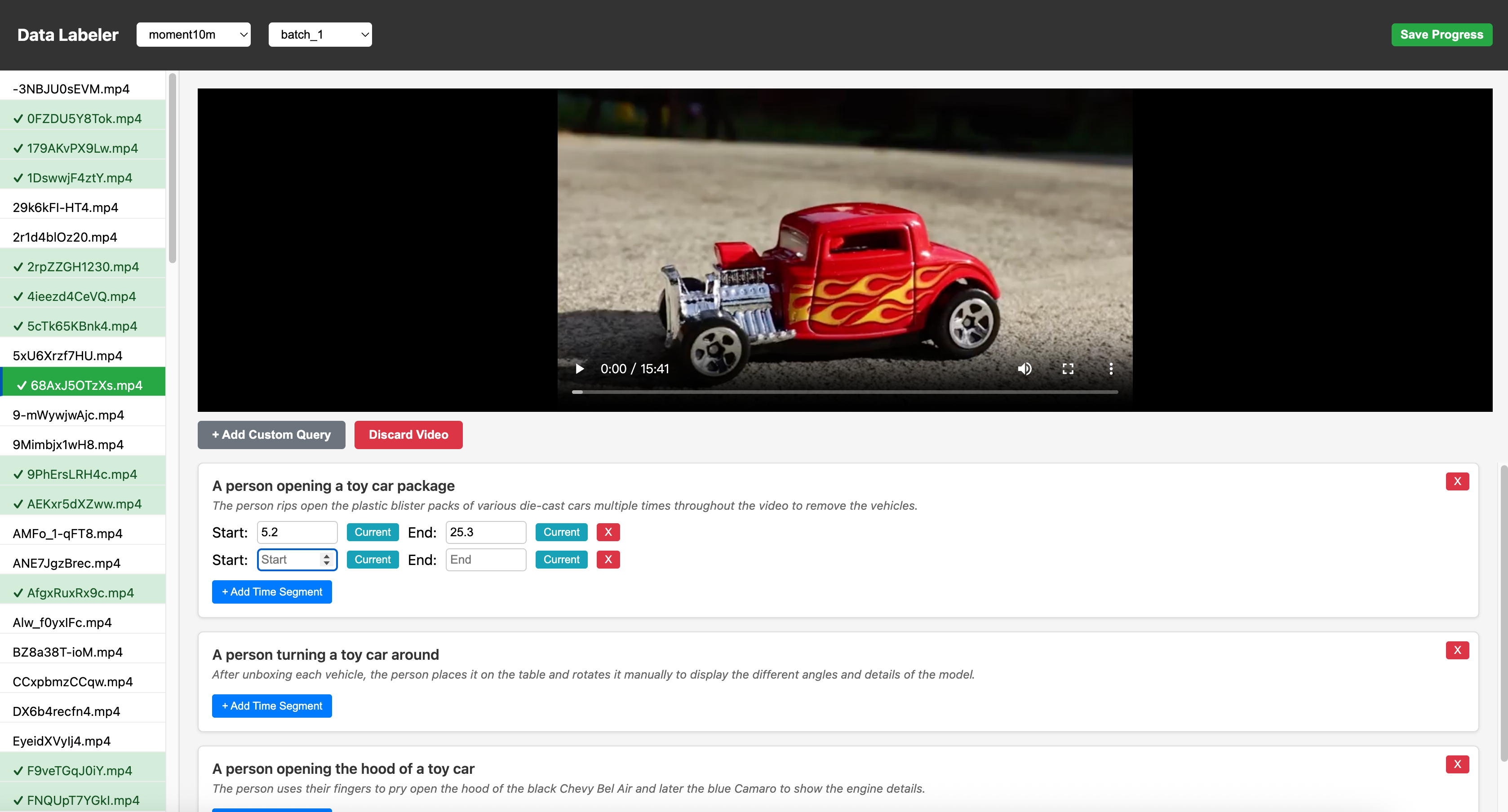}
    \caption{\textbf{Illustration of our annotation interface.} The tool allows annotators to watch videos and label multiple disjoint time segments for a given query. Pre-generated Gemini queries are provided as hints but require manual verification and adjustment.}
    \label{fig:annotation_interface}
\end{figure}

\textbf{Annotation Tool Design:} Our annotation interface is a lightweight, local web application based on Python. It is designed to handle the complexity of multi-event video retrieval, allowing annotators to mark multiple non-contiguous time segments for a single textual query. The interface, illustrated in Figure~\ref{fig:annotation_interface}.

\textbf{Annotation Workflow:} The annotation process is standardized to ensure consistency across different annotators. The workflow proceeds as follows:

\textit{1. Review and Labeling.} 
For each video, the annotator reviews the pre-generated textual queries. They watch the video to identify \textbf{all} occurrences of the described action or event. 
\begin{itemize}
    \item If the query accurately describes events in the video, the annotator marks the start and end times for \textbf{every} instance of that event.
    \item If the pre-generated query is inaccurate, the annotator modifies it.
    \item If the video contains distinct events not covered by the list, the annotator adds a new query and labels its occurrences.
\end{itemize}

\textit{2. Submission.} 
Annotations are automatically saved to a local json file. Once a batch is completed, this file is collected for quality verification.

\textbf{Annotation Quality Criteria:}
To ensure the benchmark rigorously evaluates a model's ability to handle the OMTG task, we enforcedstrict guidelines. Annotators were instructed to adhere to the following four rules:

\textit{1. Completeness (No Missing Segments):} 
    The core requirement of OMTG is to find \textit{all} instances. Annotators must ensure that every single occurrence of the query within the video is marked. Missing a segment is considered a critical error.
    
\textit{2. Temporal Tightness (Boundary Precision):} 
    Annotations must be as tight as possible. The start timestamp should mark the exact beginning of the action, and the end timestamp should mark its immediate conclusion. In cases of ambiguity (e.g., slow transitions), annotators should adopt a "conservative" approach, shrinking the window to include only the frames where the action is clearly visible, rather than including ambiguous transition frames.
    
\textit{3. Human-Perceived Clarity:} 
    Quality is prioritized over quantity. If a video is low-resolution, heavily occluded, or if the events are too ambiguous to define clearly (e.g., "someone might be smiling"), the video should be discarded using the Discard Video function. Ground truth is established solely based on clear human perception.
    
\textit{4. Preference for High Cardinality:} 
    Since existing datasets are dominated by single-segment events, our benchmark aims to fill the gap for multi-segment retrieval. Annotators are explicitly instructed to prioritize and retain queries that correspond to multiple time segments (e.g., "a person jumps" happening three times) over single-occurrence events.

\end{document}